\PassOptionsToPackage{usenames,dvipsnames,table}{xcolor}
\documentclass[letterpaper, 10 pt, conference]{ieeeconf}  

\IEEEoverridecommandlockouts                              

\usepackage{times}

\usepackage{soul,xcolor}
\usepackage{multicol}
\usepackage{graphicx}

\usepackage{amsmath,amsthm,amssymb}
\usepackage[noend]{algpseudocode}
\usepackage{amsfonts}
\usepackage{amsmath}
\usepackage{float}
\usepackage{mathrsfs}
\usepackage{booktabs}

\let\labelindent\relax

\usepackage{enumitem}
\usepackage{algorithm}
\usepackage{algpseudocode}

\usepackage{todonotes}
\usepackage{multirow}
\usepackage{stfloats}
\usepackage{etoolbox}
\usepackage{caption}
\usepackage{capt-of}
\usepackage{duckuments}
\usepackage{makecell}
\usepackage{todonotes}

\usepackage[usenames,dvipsnames,table]{xcolor}
\usepackage[bookmarks=true]{hyperref}
\hypersetup{
    colorlinks=true,
    linkcolor=MidnightBlue,
    filecolor=MidnightBlue,      
    urlcolor=MidnightBlue,
    citecolor=MidnightBlue,
}
\usepackage{cleveref}

\crefname{figure}{Fig.}{Figs.}
\Crefname{figure}{Fig.}{Figs.}
\crefname{table}{Tab.}{Tabs.}
\Crefname{table}{Tab.}{Tabs.}
\crefname{section}{Sec.}{Secs.}
\Crefname{section}{Sec.}{Secs.}
\crefname{subsection}{Sec.}{Secs.}
\Crefname{subsection}{Sec.}{Secs.}
\crefname{subsubsection}{Sec.}{Secs.}
\Crefname{subsubsection}{Sec.}{Secs.}
\crefname{appendix}{Sec.}{Secs.}
\Crefname{appendix}{Sec.}{Secs.}
\crefname{equation}{Eq.}{Eqs.}
\Crefname{equation}{Eq.}{Eqs.}
\crefname{algorithm}{Alg.}{Algs.}
\Crefname{algorithm}{Alg.}{Algs.}
\crefname{assumption}{Assumption}{Assumptions}
\Crefname{assumption}{Assumption}{Assumptions}
\crefname{proposition}{Prop.}{Props.}
\Crefname{proposition}{Prop.}{Props.}
\crefname{theorem}{Thm.}{Thms.}
\Crefname{theorem}{Thm.}{Thms.}

\theoremstyle{plain}

\theoremstyle{definition}

\title{\LARGE \bf
HULK: Learning Whole-Body Forceful Loco-Manipulation for Humanoids
}

\author{An Dang$^{*1,2}$, Arturo Flores Alvarez$^{*1,3}$, Yu-Ming Chen$^{1}$, Conor Mc Gartoll$^{1}$, Helen Sun$^{1}$, \\
Aaron Ames$^{1,4}$, Nima Fazeli$^{1,2}$,  Manikantan Nambi$^{1}$
\thanks{$^{*}$Equal contribution. This work was done during an internship at Amazon.}
\thanks{$^{1}$Amazon,
        {\tt\small <andangt,amfloalv,yumingch,cmcgtoll, sunhele,aames,nfz,mnambi>@amazon.com}}%
\thanks{$^{2}$University of Michigan,
        {\tt\small <andang,nfz>@umich.edu}}%
\thanks{$^{3}$University of California, Los Angeles,
        {\tt\small afloresa@ucla.edu}}%
\thanks{$^{4}$California Institute of Technology,
        {\tt\small ames@caltech.edu}}
}

\begin{document}

\raggedbottom

\IEEEaftertitletext{%
    \begin{center}%
        \includegraphics[width=\textwidth, height=7cm]{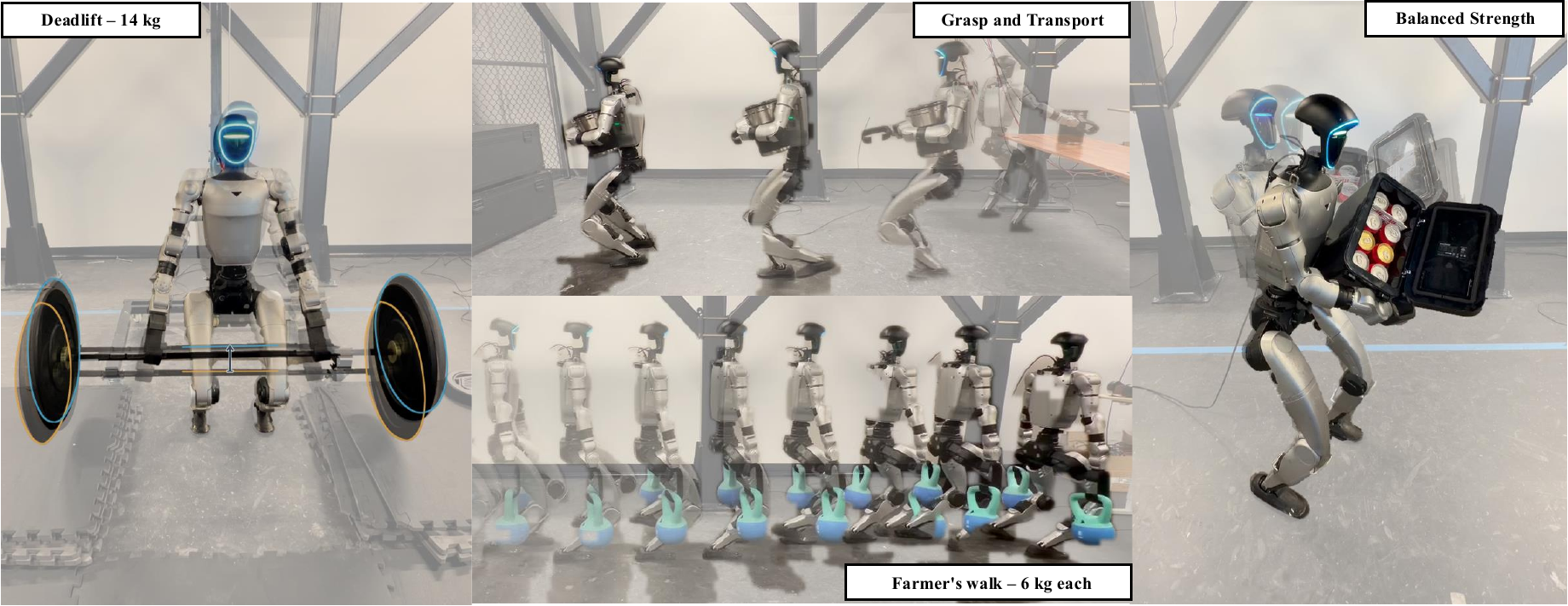}%
        \captionof{figure}{\textbf{HULK enables loco-manipulation of heavy loads, anywhere on the upper body.} HULK first trains two specialists: one that tracks arm motions under forces applied at the wrists, and one that locomotes while holding large objects against the body. Next, it distills them into a single policy that drives one whole-body command interface. A predictive model of the loaded robot guides both specialists during training, so the policy anticipates how a payload alters the robot's centroidal dynamics rather than discovering this relationship from reward alone. This pipeline enables the robot to perform on unseen tasks such as farmer's carry and deadlift. Project website: \textbf{\textcolor{MidnightBlue}{\href{http://hulk-forceful-wbc.com}{www.hulk-forceful-wbc.com}}}. }%
        \label{fig:teaser}%
    \end{center}%
}

\maketitle
\thispagestyle{empty}
\pagestyle{empty}

\begin{abstract}
Humanoid loco-manipulation of large, heavy objects demands forceful interaction across the entire body. However, such payloads shift a humanoid's center of mass and impose sustained loads across the upper body, challenging balance and tracking. We present HULK, a whole-body control framework for forceful loco-manipulation. Using model predictive control (MPC) to guide reinforcement learning with predictions of the loaded dynamics, we train two teachers: one tracks arm motions under wrist forces, and the other locomotes while holding large objects against the body. A capture-point control barrier function augments the wrist-force teacher during training to improve balance under load. We distill both teachers into a single policy. We evaluate our method on a Unitree G1 in both simulation and the real-world. In simulation, the teacher with the barrier function achieves the lowest forward and lateral velocity tracking errors at $10$~kg per arm among evaluated controllers and reduces aggregate balance metric by $35.7\%$ relative to MPC-guided reinforcement learning alone. Our wrist-force teacher withstands torso push disturbances of up to $130$~N.
\end{abstract}

\section{Introduction}

Humanoid robots are a compelling form factor for a world
built around the human body. The promise is a single platform that can
perform the same physical labor as people, in the same unmodified
spaces. Realizing this promise, however, requires forceful
whole-body work: lifting a loaded tote, leaning into a cart, or
carrying a heavy object against the chest while walking. These tasks recur throughout logistics, manufacturing, and assistive
applications. What these tasks share is a load the robot must
sustain rather than reject: it persists for the task's duration, shifts where the robot's weight sits, and competes for
actuation that keeps it balanced.

Current approaches to humanoid loco-manipulation have converged on
a layered control strategy. At the bottom sits a learned whole-body
controller (WBC): a policy that turns a compact set of task-space
commands, such as base velocity, torso pose, and arm targets, into
joint actions that keep the robot balanced. Above it, planners,
teleoperation interfaces, and agentic pipelines issue those
commands without reasoning about balance themselves, treating the
WBC as a low-level ``System~0''~\cite{handoff2026, psi02026}.
Massively parallel reinforcement learning (RL) in simulation has
made this bottom layer reliable enough to serve as that
substrate~\cite{omnih2o2024, hover2025, exbody2024, humanplus2024,
sonic2025, beyondmimic2025}, and progress above it has been rapid
along two fronts: \emph{agility}, pushing the dynamism of
whole-body skills~\cite{asap2025, xloco2026, omnixtreme2026}, and
\emph{perception}, closing the loop on terrain and
obstacles~\cite{rpl2026, php2026}.

Forceful interaction, however, remains comparatively underexplored. Unlike agility and perception, it requires the WBC to adapt its balance strategy to large and sustained external wrenches. A heavy payload changes the forces and moments that must be supported by the robot, with their effect varying continuously with the arm configuration and the object's pose relative to the body. Existing approaches address parts of this problem. Compliance-based methods lower the effective stiffness of the robot so that it yields under unexpected contact~\cite{softmimic2025}; while effective for incidental interaction, this behavior is poorly suited to supporting a heavy payload against gravity. The strongest forceful controllers instead learn to resist applied forces, but only \emph{at the wrists}~\cite{falcon2026}, or only for horizontal push--pull moments~\cite{thor2025, hafo2025}. Carrying a load further requires this adaptation to persist: the disturbance is not a transient to be rejected, but a condition that remains throughout the task. The robot must therefore maintain balance under a continuously applied wrench while simultaneously producing the forces required for manipulation.

\begin{figure*}[t]
    \centering
    \includegraphics[trim=10.5bp 183bp 37.25bp 182bp, clip,
                     width=\textwidth]{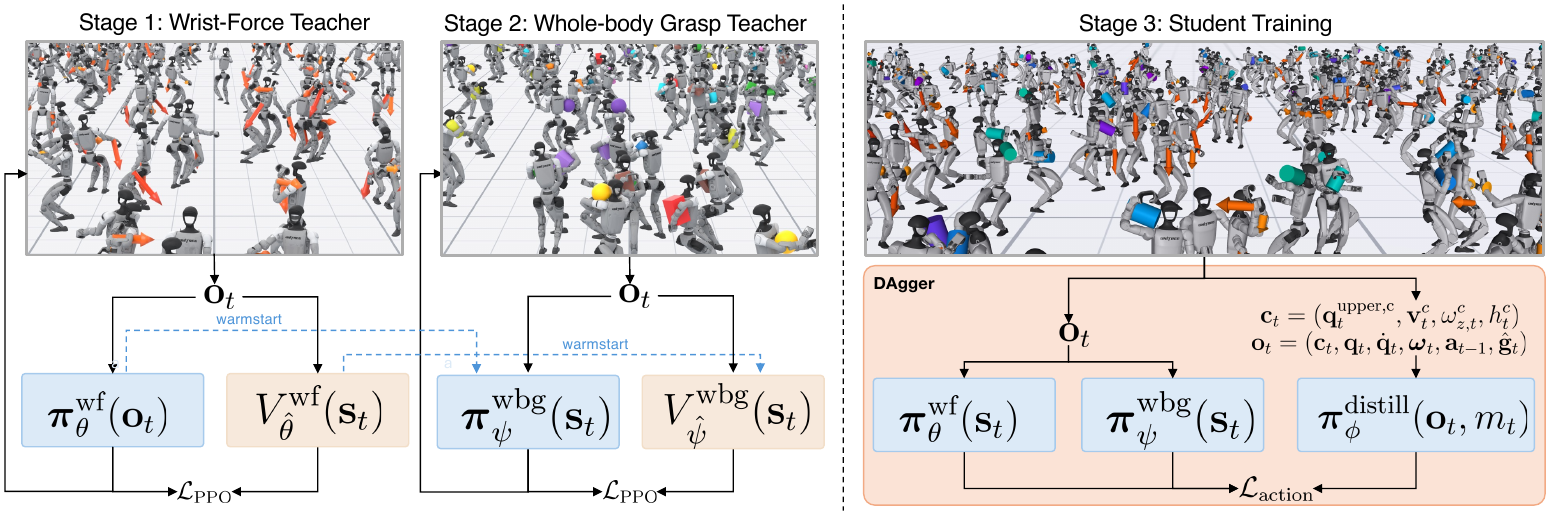}
    \caption{\textbf{Training overview.} We decompose training into 3 stages. In stage 1, we train a teacher $\boldsymbol{\pi}_\theta^{\text{wf}}$ that follows high-level commands $\mathbf{c}_t$ while undergoing force perturbations at the wrists. High-level commands include upper body joint targets $\mathbf{q}^{\text{upper},c}_t$, commanded linear velocity $\mathbf{v}_t^c$, commanded angular velocity in z direction $\boldsymbol{\omega}_t^c$, and height command $h^c_t$. In stage 2, the teacher $\boldsymbol{\pi}_\psi^{\text{wbg}}$ follows these commands while maintaining whole-body grasps across multiple object geometries. In stage 3, the student policy $\boldsymbol{\pi}_\phi^{\text{distill}}$ is trained using DAgger with teacher mode conditioning $m_t \in \{0,1\}$. The student takes in sensor observations $\mathbf{o}_t$ only present on a humanoid such as high-level commands, joint position $\mathbf{q}_t$, joint velocity $\dot{\mathbf{q}}_t$, angular velocity $\boldsymbol{\omega}_t$, previous action $\mathbf{a}_{t-1}$, and projected gravity $\hat{\mathbf{g}}_t$.}
    \label{fig:training-overview}
    \vspace{-6pt}
\end{figure*}

We address the displaced center of mass and sustained balance under
load in a single learned policy. Model predictive
control (MPC) supplies reward guidance during RL~\cite{mpcrl2026},
giving the policy a model-based signal for how a load moves the
centroidal state rather than requiring it to discover that
relationship from reward alone. We then augment the wrist-force teacher
with a capture-point control barrier function (CBF)~\cite{cbfqp2017, cbftheory2019}.
Following CBF-RL~\cite{cbfrl2026}, the barrier acts during training to improve
learned balance recovery under load while retaining network-only inference. Our contributions are:


\begin{enumerate}
    \item We extend learned whole-body control to forceful interactions
    between external objects and contacts across the humanoid upper body.
    \item We use MPC reward guidance to train robust locomotion and
    loco-manipulation under sustained payloads.
    \item We augment the wrist-force teacher with training-time
    capture-point constraints, improving loaded tracking and balance
    recovery while retaining network-only inference.
\end{enumerate}


\section{Related Works}
\label{sec:related}

\begin{figure*}[t]
    \centering
    \includegraphics[trim=137.92bp 223.84bp 108.64bp 208.96bp, clip,
                     width=\textwidth]{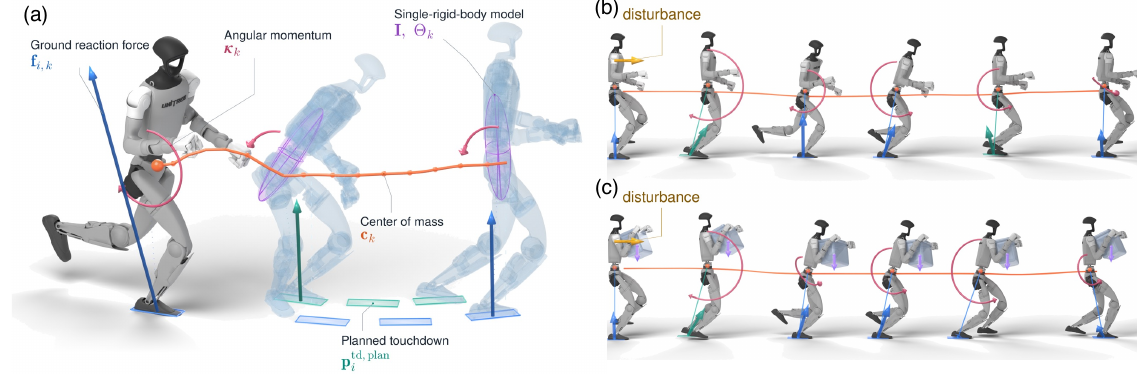}
    \caption{\textbf{Centroidal MPC Force Reasoning.} In (a), we show the G1 about to fall over with forward angular momentum. The MPC generates a plan of the CoM trajectory and ground reaction forces to re-stabilize the angular momentum of the robot and keep an upright posture. The green annotations correspond to left foot while the blue ones correspond to right foot. The light blue ghosts illustrate the G1 following the MPC plan. (b) and (c) showcase the G1 stabilizing from an external pushes in a hands-free and minifridge grasp scenario.}         
    \label{fig:mpc-force-reasoning}
    \vspace{-0pt}
\end{figure*}

\noindent \textbf{Whole-body Control.}
Whole-body controllers have lately been trained using reinforcement learning to follow high-level commands while actively stabilizing the humanoid. Several works \cite{omnih2o2024, hover2025, exbody2024, humanplus2024, beyondmimic2025} demonstrate the capabilities of sim-to-real reinforcement learning for motion tracking or teleoperation of the humanoid for human-like behavior on hardware. Large-scale models now expose whole-body control as a general-purpose interface: SONIC~\cite{sonic2025} supersizes motion tracking toward natural whole-body control using a large human motion tracking dataset. These methods address kinematic tracking rather than forces and persistent loading.

\vspace{3pt}
\noindent \textbf{Forceful Whole-body Control.}
Two efforts define the forceful direction. FALCON~\cite{falcon2026} rejects force perturbations at the wrists with a wrist-force curriculum and a dual-network architecture, but struggles to hold commands while doing so. Thor~\cite{thor2025} leans the torso to recruit the robot's own weight, using ZMP analysis of sagittal-plane wrist forces, and so addresses horizontal push--pull rather than downward payload forces. Neither models whole-body object interaction during training, so the payload-augmented center of mass degrades locomotion under heavy load. HAFO~\cite{hafo2025} and load-aware adaptation~\cite{kang2026splitadapterloadawarehumanoidlocomanipulation} target force-adaptive control, but heavy payloads remain open, while Hold~My~Beer~\cite{holdmybeer} and FACET~\cite{facet} stabilize a payload under gentle motion rather than sustaining large loads.

Outside the learned-WBC setting, payload carrying has been addressed by hierarchical optimization~\cite{rigo2024heavyobjects} and by distilling trajectory optimization into skills~\cite{opt2skill}, both indicating that a model of the loaded dynamics is valuable; Dao~et~al.~\cite{dao2024boxlocomanipulation} show sim-to-real box loco-manipulation without a whole-body command interface. AMO~\cite{amo} provides such an interface but assumes both feet remain in contact.

Closest to our approach, Li~et~al.~\cite{mpcrl2026} use model predictive control for reward guidance and demonstrates cart-pushing and payload carrying. However, this method treats interactions as forces applied at the wrists rather than incorporating a payload wrench into reference generation, and it does not address contact against the torso. We extend it to a whole-body controller accepting height, torso-tilt and arm commands, with payloads entering the model through force perturbations and whole-body grasp force profiles.

\vspace{3pt}
\noindent\textbf{Balance Constraints in Learned Control.} Control barrier functions are ordinarily enforced by a runtime quadratic program that minimally corrects a nominal controller~\cite{cbfqp2017, cbftheory2019, taylor2020learningcbf}, which for a learned policy means shipping a reduced-order model, gradient and solver alongside the network. CBF-RL~\cite{cbfrl2026} instead applies the correction during training and penalizes the residual in the reward, so the requirement is internalized and deployment carries no extra machinery. We adopt that formulation. Because a sustained external wrench displaces the center of mass for a whole episode, the quantity to constrain is a balance state under load, for which the capture point model~\cite{pratt2006capture}, is the natural choice.






\section{Methodology}



Our goal is to learn a single whole-body controller for payload-driven loco-manipulation: a policy that follows planar velocity, base height, torso orientation and arm joint commands while a load acts anywhere on the upper body, and that transfers to hardware from proprioception alone. We train two specialist teachers separately, one that tracks arm poses under forces applied at the wrists and one that locomotes while maintaining whole-body grasps across multiple objects. We distill the two into the single student we deploy (\cref{sec:distill}, \cref{fig:training-overview}). The teachers both share a reward whose force-aware terms are supplied by a predictive model of the loaded robot rather than hand-designed (\cref{sec:guidance}), using a centroidal model as the predictor (\cref{sec:dynamics}) together with the references that inform it of the load (\cref{sec:footstep}), and balance imposed by a barrier function during training instead of a runtime filter (\cref{ssec:cbf}). The two differ only in the interaction each is exposed to and how it enters the model, which we make concrete once per teacher in \cref{sec:teachers}.

\subsection{Problem Formulation}

We formulate forceful whole-body control as an infinite-horizon discrete-time partially observable Markov decision process (POMDP) defined by the tuple $\mathcal{M} := (\mathcal{S}, \mathcal{O}, \mathcal{A}, \mathcal{R}, \mathcal{P}, \rho_0, \gamma)$ with state space $\mathcal{S}$, observation space $\mathcal{O}$, action space $\mathcal{A}$, reward $\mathcal{R}(\mathbf{s}, \mathbf{a})$, transition probability $\mathcal{P}(\mathbf{s}' \mid \mathbf{s}, \mathbf{a})$, initial state distribution $\rho_0$, and discount factor $\gamma \in [0,1)$. The robot seeks a policy $\pi_\theta(\mathbf{a} \mid \mathbf{o})$, parameterized by $\theta$, that maximizes the expected discounted return $\mathcal{J} = \mathbb{E}_{\pi_\theta}\big[ \sum_{t=0}^{\infty} \gamma^{t} r_t \big]$ over control steps $t$, where $\mathbf{s} \in \mathcal{S}$, $\mathbf{o} \in \mathcal{O}$, $\mathbf{a} \in \mathcal{A}$, and $r_t = \mathcal{R}(\mathbf{s}_t, \mathbf{a}_t)$.

\subsection{MPC Reward Guidance}
\label{sec:guidance}

We use MPC to generate outputs that will be tracked in the reward formulation of both teachers. These outputs include contact/interaction forces between the robot and its environment, the robot center of mass (CoM), and the rotation of the robot pelvis. We use MPC to solve the following optimization problem over a horizon of $N$ steps of duration $\Delta t$, given a dynamics model and constraints:
\begin{subequations}
\label{eq:mpc}
\begin{align}
    \min_{\mathcal{Z}} \quad & \sum_{k=0}^{N-1} L_k(\mathbf{x}_k, \mathbf{u}_k) + L_{\Delta,k}(\mathbf{u}_k, \mathbf{u}_{k-1}) + L_N(\mathbf{x}_N) \label{eq:mpc-cost}\\
    \text{s.t.} \quad & \mathbf{x}_{k+1} = f_{\Delta t}(\mathbf{x}_k, \mathbf{u}_k), \quad k = 0, \dots, N-1, \label{eq:mpc-dyn}\\
    & \mathbf{u}_k \in \mathcal{U}_k, \quad k = 0, \dots, N-1, \label{eq:mpc-con}\\
    & \mathbf{x}_0 = \Phi(\mathbf{q}_t, \dot{\mathbf{q}}_t), \label{eq:mpc-init}
\end{align}
\end{subequations}
where $\mathcal{Z} = \{ \mathbf{x}_1, \dots, \mathbf{x}_N, \mathbf{u}_0, \dots, \mathbf{u}_{N-1} \}$, $\mathbf{x}_k$ and $\mathbf{u}_k$ are the state and control input at horizon step $k$ (both made concrete in \cref{sec:dynamics}), $f_{\Delta t}$ is the dynamics model discretized at $\Delta t$, $\Phi$ maps the measured full-order configuration and velocity $(\mathbf{q}_t, \dot{\mathbf{q}}_t)$ to the reduced-order state, and $\mathcal{U}_k$ represents the control input set that each $\mathbf{u}_k$ is an element of. We define our stage and terminal costs as quadratic costs defined as:
\begin{align}
    L_k &= \| \mathbf{x}_k - \mathbf{x}^{\mathrm{ref}}_k \|^2_{\mathbf{Q}} + \| \mathbf{u}_k - \mathbf{u}^{\mathrm{ref}}_k \|^2_{\mathbf{W}}, \label{eq:stagecost}\\
    L_{\Delta,k} &= \|\mathbf{u}_k-\mathbf{u}_{k-1}\|^2_{\mathbf{W}_{\Delta}} \\
    L_N &= \| \mathbf{x}_N - \mathbf{x}^{\mathrm{ref}}_N \|^2_{\mathbf{Q}_N}, \label{eq:termcost}
\end{align}
where $\| \mathbf{v} \|^2_{\mathbf{A}} = \mathbf{v}^{\top} \mathbf{A} \mathbf{v}$ for any vector $\mathbf{v}$ and weight matrix $\mathbf{A}$, $\mathbf{x}^{\mathrm{ref}}_k$ is the reference trajectory generated from the high-level commands, $\mathbf{u}^{\mathrm{ref}}_k$ is an input reference that is zero unless a known external load is being planned for (\cref{sec:teachers}), so that the second term is plain input regularization by default, $\mathbf{u}_{-1}$ is the input applied at the previous solve, and $\mathbf{Q}$, $\mathbf{W}$, and $\mathbf{Q}_N$ represent tunable diagonal positive semi-definite matrices for the MPC optimization problem.

Solving~\eqref{eq:mpc} yields an optimal state trajectory $\{ \tilde{\mathbf{x}}_k \}_{k=0}^{N}$ and input trajectory $\{ \tilde{\mathbf{u}}_k \}_{k=0}^{N-1}$. Rather than executing the plan, we use it as reward guidance for RL~\cite{mpcrl2026}: the policy is rewarded for keeping the measured robot state close to the plan,
\begin{equation}
    r^{\mathrm{mpc}}_t = \sum_{i \in \mathcal{T}} \exp\!\big( -\alpha_i\, \| \Lambda_i(\tilde{\mathbf{x}}_{k(t)}, \tilde{\mathbf{u}}_{k(t)}) - \Lambda_i(\mathbf{x}_t, \mathbf{u}_t) \|^2 \big),
    \label{eq:mpcreward}
\end{equation}
where $\mathcal{T}$ indexes the tracked quantities, each $\Lambda_i$ extracts one of them from the plan at the horizon step $k(t)$ corresponding to time $t$ and from the measured reduced-order state $\mathbf{x}_t$ and measured interaction forces $\mathbf{u}_t$, and $\alpha_i > 0$ sets the sensitivity of the $i$-th term; the tracked quantities themselves follow from the choice of model (\cref{sec:dynamics}). This gives the policy a signal for how the payload displaces the reduced-order state, rather than requiring it to discover the loaded dynamics from task rewards alone.

\subsection{Centroidal Dynamics}
\label{sec:dynamics}

For reward guidance the plan need only be dynamically consistent rather than full-order, so we adopt the reduced-order model of~\cite{mpcrl2026} and extend its interaction set to the upper body. The state of~\eqref{eq:mpc} is the centroidal state $\mathbf{x}_k = (\mathbf{c}_k, \boldsymbol{\Theta}_k, \mathbf{l}_k, \boldsymbol{\kappa}_k) \in \mathbb{R}^{12}$, collecting the CoM position, the base roll, pitch and yaw, and the centroidal linear and angular momenta, respectively. The input $\mathbf{u}_k = (\mathbf{f}_{i,k}, \boldsymbol{\tau}_{i,k})_{i \in \mathcal{C}}$ stacks the wrenches at the interaction points $\mathcal{C}$, the two feet and, when they carry load, the two wrists; a flat foot contributes a six-dimensional wrench and a wrist a three-dimensional force, so $\mathbf{u}_k \in \mathbb{R}^{6|\mathcal{C}|}$. With $m$ the total mass, $\mathbf{g}$ gravity, $\mathbf{I}$ the composite rigid-body inertia held constant over the horizon, $\sigma_{i,k} \in \{0,1\}$ gating interaction $i$ into the active set $\mathcal{C}_k$, and $\bar{\mathbf{c}}_k$ a reference CoM, an Euler discretization of the centroidal dynamics~\cite{orin2013centroidal} gives
\begin{equation}
\begin{aligned}
    \mathbf{c}_{k+1} &= \mathbf{c}_k + \tfrac{\Delta t}{m} \mathbf{l}_k + \tfrac{\Delta t^2}{2} \Big( \mathbf{g} + \tfrac{1}{m} \textstyle\sum_{i \in \mathcal{C}} \sigma_{i,k} \mathbf{f}_{i,k} \Big), \\
    \boldsymbol{\Theta}_{k+1} &= \boldsymbol{\Theta}_k + \Delta t\, \mathbf{I}^{-1} \boldsymbol{\kappa}_k, \\
    \mathbf{l}_{k+1} &= \mathbf{l}_k + \Delta t \Big( m \mathbf{g} + \textstyle\sum_{i \in \mathcal{C}} \sigma_{i,k} \mathbf{f}_{i,k} \Big), \\
    \boldsymbol{\kappa}_{k+1} &= \boldsymbol{\kappa}_k + \Delta t \textstyle\sum_{i \in \mathcal{C}} \sigma_{i,k} \big[ (\mathbf{p}_{i,k} - \bar{\mathbf{c}}_k) \times \mathbf{f}_{i,k} + \boldsymbol{\tau}_{i,k} \big],
\end{aligned}
\label{eq:affine}
\end{equation}
which is affine in $(\mathbf{x}_k, \mathbf{u}_k)$, hence a linear time-varying model $\mathbf{x}_{k+1} = \mathbf{A}_k \mathbf{x}_k + \mathbf{B}_k \mathbf{u}_k + \mathbf{e}_k$. A carried payload is not a decision variable: the force and moment it exerts about the CoM enter $\mathbf{e}_k$ in the same rows as an interaction wrench, so the plan leans into the load rather than rejecting it as a disturbance.

The input set $\mathcal{U}_k$ of~\eqref{eq:mpc-con} collects the wrenches an active interaction can source. A contact force obeys Coulomb friction, $\|\mathbf{f}^t_i\|_2 \leq \mu_i f^n_i$ with $f^n_i \in [0, f^{n,\max}_i]$ for tangential and normal components $\mathbf{f}^t_i, f^n_i$ and friction coefficient $\mu_i$; bounding the two tangential components individually replaces the second-order cone by halfspaces, keeping the problem a quadratic program. Every row scales with $f^n_i$, so a gated swing foot can source no force.

\subsection{Force-Aware References}
\label{sec:footstep}
The centroidal MPC prescribes footstep locations and gait timings to preserve convexity. Prior work, however, does not account for sustained external forces when generating these footstep references. The footstep planner used in Li~et~al.~\cite{mpcrl2026} uses a feedforward Raibert heuristic~\cite{raibert1986} that places the swing foot at the neutral point of the commanded stride,
\begin{equation}
    \mathbf{p}^{\mathrm{td}}_i = \mathbf{c}^{\mathrm{td}}_{xy} + \frac{T_{\mathrm{st}}}{2}\, \mathbf{v}^{\mathrm{cmd}}_{xy} + \mathbf{R}_z(\psi)\, \mathbf{d}_i,
    \label{eq:raibert}
\end{equation}
where $\mathbf{p}^{\mathrm{td}}_i \in \mathbb{R}^2$ is the touchdown location of foot $i$, the subscript $xy$ denotes planar components, $T_{\mathrm{st}}$ is the stance duration, $\mathbf{v}^{\mathrm{cmd}}_{xy}$ the commanded planar velocity, $\mathbf{R}_z(\psi)$ the rotation by the base yaw $\psi$, $\mathbf{d}_i$ the lateral offset of foot $i$ from the CoM in the base frame, and $\mathbf{c}^{\mathrm{td}}_{xy}$ the CoM position predicted at touchdown by integrating the commanded twist forward from the measured $\mathbf{c}_{xy}$. Equation~\eqref{eq:raibert} is purely feedforward. Under a sustained external force the realized velocity deviates from $\mathbf{v}^{\mathrm{cmd}}_{xy}$.

We make the Raibert heuristic force-aware by adding an offset for external
force perturbations. Let $\mathcal{H}$ be the set of contacts on the robot's
body, where contact $j$ applies a force $\mathbf{f}^{e}_{j,xy}$ at height
$z_j$ above the ground. Balancing moments about the CoM, a foot displacement $d$ contributes $mg\,d$
while a force $\mathbf{f}^{e}_{j,xy}$ applied at height $z_j$ contributes
$z_j \mathbf{f}^{e}_{j,xy}$. Equating the two, the foot must land further out
by $\sum_{j \in \mathcal{H}} z_j \mathbf{f}^{e}_{j,xy} / (mg)$. Adding this
offset to Eq.~\eqref{eq:raibert} and scaling it by a gain $k_f$ gives
\begin{align}
    \mathbf{p}^{\mathrm{tdf}}_i = \mathbf{p}_i^{td} + \frac{k_f}{m g} \sum_{j \in \mathcal{H}} z_j\, \mathbf{f}^{e}_{j,xy},
    \label{eq:forceraibert}
\end{align}
We also add the Raibert feedback term~\cite{raibert1986} where $k_c$ is the gain controlling feedback law which results in $\mathbf{p}^{\mathrm{tdf-fb}}_i = \mathbf{p}^{\mathrm{tdf}}_i + \frac{k_c}{\omega_0} \big( \dot{\mathbf{c}}_{xy} - \mathbf{v}^{\mathrm{cmd}}_{xy} \big) \nonumber$.

\subsection{Teacher Training}
\label{sec:teachers}

\textbf{Wrist-force teacher.} We train a controller that braces and resists against forces applied at the wrists while tracking an upper-body pose command, using a force curriculum akin to \cite{falcon2026}. The arm command is drawn from the BONES-RP1 dataset \cite{bones-rp1} retargeted to the $29$-DoF robot: we keep the $14$ arm joint angles of every retargeted frame, roughly $4.8 \times 10^4$ frames over $252$ motions, and sample individual frames as static pose targets rather than replaying a trajectory, so the teacher has to hold arbitrary arm configurations under load instead of one scripted motion. Independent 3D forces are applied at the two wrists, up to $\pm 40$\,N horizontally and $-50$ to $+5$\,N vertically, each axis zeroed with probability $0.25$, held for $3$ to $5$\,s, and ramped by a per-environment magnitude curriculum. The wrists enter the model~\eqref{eq:affine} as two point interactions alongside the feet, placed by forward kinematics along the arm chain evaluated at the \emph{commanded} arm target and gated on only while a force is live, so $\mathbf{u}_k \in \mathbb{R}^{18}$ collects two foot wrenches and two wrist forces.

\textbf{Whole-body-grasp teacher.} The teacher learns to locomote and stay stable while maintaining whole-body grasps across multiple objects. The interactions are the ground reaction forces and the contact forces of each environment's assigned grasp. We generate 130,000 grasps offline across multiple object geometries using energy-based optimization from \cite{ye2025dex1blearning1bdemonstrations} (examples in \cref{fig:app-grasp-examples}). Each environment is initialized with an object and a compatible grasp to maintain, so the contact locations of~\eqref{eq:affine} are read from that grasp and gated on for as long as the assigned object is held. The grasp contact forces are decision variables subject to the same
friction constraints as the ground-reaction forces. The forces exerted
by the object in the sampled whole-body grasp provide
$\mathbf{u}^{\mathrm{ref}}_k$, while the MPC optimizes the corresponding
interaction forces. These forces also contribute to the centroidal
angular-momentum dynamics in \cref{eq:affine}. All else in \crefrange{sec:guidance}{sec:footstep} is shared between the two teachers. During training, we warm-start from the wrist-force teacher so that locomotion and upper-body command tracking are not relearned from scratch.


\begin{figure}[t]
    \centering
    \includegraphics[width=\columnwidth]{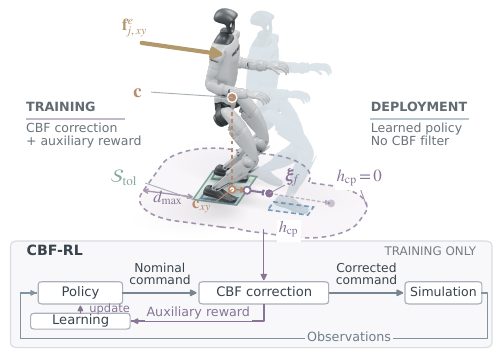}
    \caption{\textbf{CP-CBF guidance.} The ghost illustrates a recovery step into a separate admissible footprint; the dashed boundary marks $h_{\mathrm{cp}}=0$. The inset shows command correction and reward feedback during training only. }
    \label{fig:cp-cbf-guidance}
    \vspace{-6pt}
\end{figure}

\subsection{Capture-Point Balance Guidance via CBF-RL}
\label{ssec:cbf}

Forceful loco-manipulation demands sustained balance while the robot keeps tracking motion and force commands. We introduce a capture-point control barrier function (CP-CBF) and incorporate it into the wrist-force teacher through CBF-RL~\cite{cbfrl2026}. This corrects the wrist-force teacher's commands with a control barrier function during training and penalizes the residual in the reward, so the exported policy runs unfiltered (\cref{fig:cp-cbf-guidance}). We defer to~\cite{cbfrl2026} for the training mechanics.

The barrier uses the instantaneous capture point~\cite{pratt2006capture}, equivalently the planar divergent component of motion (DCM) under the constant-height LIPM, defined as $\boldsymbol{\xi} = \mathbf{c}_{xy} + \dot{\mathbf{c}}_{xy}/\omega_0$, the support location that would arrest horizontal motion under a linear inverted pendulum. Here $\mathbf{c}_{xy}$ and $\dot{\mathbf{c}}_{xy}$ are the horizontal CoM position and velocity, and $\omega_0=\sqrt{g/z_{\mathrm{cmd}}}$ is the natural frequency with $z_{\mathrm{cmd}}$ the CoM height from the commanded base height.
We make the barrier force-aware as $\boldsymbol{\xi}_f=\boldsymbol{\xi}+\Delta\mathbf{p}_{\mathrm{CoP}}$, where $\Delta\mathbf{p}_{\mathrm{CoP}}$ is the center-of-pressure shift that balances the external wrench. We then measure $\boldsymbol{\xi}_f$ against a support region $\mathcal{S}_{\mathrm{tol}}$, a yaw-aligned union of tolerance-expanded boxes over the contacting soles and predicted touchdown footprints that admits recovery steps without covering the intervening region, and define
\begin{equation}
    h_{\mathrm{cp}} = 1 - \operatorname{dist}(\boldsymbol{\xi}_f,\mathcal{S}_{\mathrm{tol}})/d_{\max},
    \label{eq:hcp}
\end{equation}
where $\operatorname{dist}$ is the minimum distance to the region, zero inside it, and $d_{\max}>0$ is the admissible excursion, so that $h_{\mathrm{cp}}>0$ is remaining margin, $h_{\mathrm{cp}}=0$ its limit, and negative values an exceeded allowance. The admissible excursion we denote as the divergent component of motion (DCM) excursion. This will also be the balance metric that we use in our experiments.
The CBF-QP and reward guidance are detailed in \cref{app:cbf-training}.

\subsection{Student Distillation}
\label{sec:distill}

The teachers specialize in different interaction regimes. We freeze both and train a mode-conditioned student with DAgger~\cite{dagger2011}. At each student-visited state, the teacher selected by $m_t \in \{0,1\}$ supplies the target action. The student regresses these actions using the same observation space $\mathcal{O}$ as the teacher actors, keeping supervision on its own rollout distribution.



\section{Experiments}

We evaluate whether HULK can maintain commanded motion under sustained load,
recover from disturbances, and transfer these capabilities to forceful
loco-manipulation tasks.

\subsection{Experimental Setup}
\label{sec:exp-setup}

\textbf{Controllers and comparisons.} We compare against FALCON, Thor, and
AMO~\cite{falcon2026,thor2025,amo}. The simulation rows \emph{Ours}
and \emph{Ours + CP-CBF} evaluate the wrist-force teachers; hardware trials evaluate the distilled student deployed zero-shot on the Unitree~G1.

In simulation, all controllers receive identical commands and perturbation
schedules on the same platform. Commands use their shared interface:
$v_x,v_y \in [-0.5,0.5]$~m/s, $\omega_z \in [-0.5,0.5]$~rad/s,
base height $0.5$--$0.8$~m, torso roll $\pm0.2$~rad, torso pitch
$\pm0.25$~rad, and eight common arm joints. We report mean $\pm$ sample
standard deviation across five evaluation-seed aggregates for each fixed
policy. These seeds vary evaluation inputs, not policy training.
Bold marks the best displayed value in each simulation-table column.

\textbf{Carrying and push recovery.} We apply $10$~kg per arm at
mid-forearm, ramp the load over $2$~s, and allow another $2$~s to settle
before issuing commands. Each trial commands one velocity axis, measuring
tracking under sustained load after the loading transient.
Separately, standing push tests apply a one-second torso force at six
magnitudes from $50$ to $130$~N, from behind, in front, and laterally.
Each HULK teacher is evaluated on $90$ push episodes.
We measure the signed margin of the DCM defined in Section~\ref{ssec:cbf}
relative to the benchmark's two-foot support envelope.
Negative margins indicate that the DCM lies outside
this envelope; we call these departures DCM excursions. We report the
average of each episode's minimum post-push margin, with less negative
values indicating smaller worst departures outside the envelope.
Peak torso inclination is averaged over the three directions at $50$~N,
the largest push all controllers survive. Maximum push denotes the
largest tested magnitude survived in every direction.

\textbf{Wrist-force rejection.} While the robot follows arm and velocity
commands, we apply randomized 3D wrist forces with scales
$\alpha_g \in \{0,0.5,1.0\}$, corresponding to $0$, $20$, and $40$~N
per wrist. The planar component opposes travel in the base
frame. Upper-body error $E_{\mathrm{upper}}$ is the RMS tracking error
over the common arm joints. Composite root error $E_{\mathrm{root}}$
is the time average of the numerical sum of horizontal velocity-error
magnitude (m/s) and absolute yaw-rate error (rad/s); it is not a
single-unit physical quantity.

\begin{figure}[!t]
    \centering
    \includegraphics[width=0.86\columnwidth,keepaspectratio]{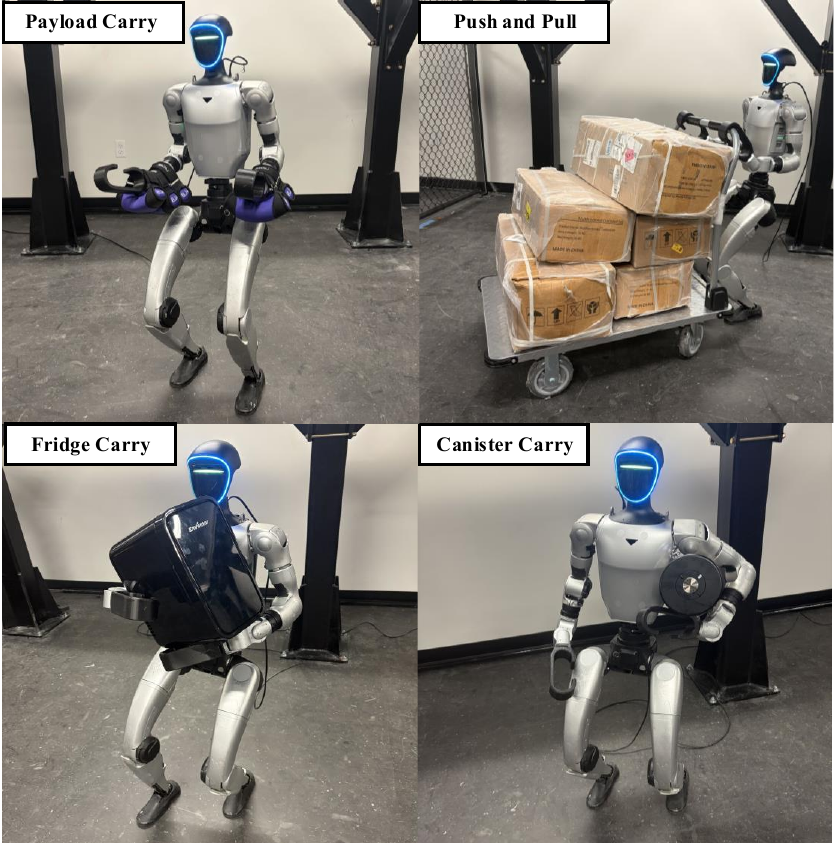}
    \caption{\textbf{Real-world tasks.} Cart push/pull, payload carry, fridge carry, and
canister carry on the Unitree~G1. Together they span the upper body: force acts at the
wrists in the first two and against the torso in the last two, and every task is performed
while walking.}
    \label{fig:sim-tasks}
    \label{fig:setup}
    \vspace{-6pt}
\end{figure}

\textbf{Hardware tasks.} Cart push/pull and payload carry load the
wrists, while fridge and canister carry additionally introduce contact
against the torso (\cref{fig:setup}). All tasks involve walking. For each
task, controllers receive identical commands and object placement at a
common load. A trial succeeds if the robot completes the commanded task
without falling, triggering a joint-torque protective stop, or dropping
the object. For payload carry, where all controllers can complete the
carrying task, we additionally require the robot to remain within a
prescribed path-deviation threshold, measured using an OptiTrack motion
capture system. Trial counts vary by task and appear as the denominators
in \cref{tab:real}.

For each controller and task, we additionally sweep the payload magnitude
to determine the maximum load completed successfully. Carrying tasks use
forward locomotion, whereas cart push/pull uses backward locomotion.
Success rates are evaluated separately at a common fixed load for each
task: 150~kg for cart push/pull, 11~kg for payload carry, 6~kg for fridge
carry, and 5~kg for canister carry.


\subsection{Simulation Results}
\label{sec:sim-eval}

\textbf{Sustained load and recovery quality.} The CP-CBF teacher combines
lower loaded translational errors with smaller DCM excursions
(\cref{tab:carrying_load}). Relative to \emph{Ours}, the largest tracking
improvement is lateral, suggesting that the benefit depends on the
commanded direction. Because the load acts at the forearms rather than
the training wrists, these results also show transfer across force
application points. AMO retains the lowest mean turn-rate error, so the
advantage does not extend uniformly across commands. The two HULK teachers achieve similar push-survival performance, but differ substantially in DCM excursion. The smaller DCM excursions of the CP-CBF teacher suggest improved recovery quality that push survival alone may hide. The smaller DCM excursions from our CP-CBF teacher suggest a difference in recovery quality that push survival may hide. This is consistent with the balance state targeted during training, although negative margins show that recovery involves leaving the support envelope.

\textbf{Arm accuracy and locomotion under wrist forces.}
Both HULK teachers perform competitively against FALCON in arm and
composite root tracking at the strongest tested wrist force
(\cref{tab:hand_rejection_walking}) with only a mean difference of $0.02$~rad. Their arm errors remain nearly constant as force increases,
although FALCON retains the lowest means. CP-CBF has higher root error
than our baselines at both nonzero force levels. Our teachers train on sampled poses held under load, with MPC and CP-CBF guidance emphasizing balance, while FALCON uses a torque-aware force curriculum~\cite{falcon2026}.

\subsection{Real-world Results}
\label{sec:results}


HULK achieves the highest observed completion rate and maximum completed
load on three of the four hardware tasks, and ties AMO on payload carry
(\cref{tab:real}). In cart push/pull, HULK completes $5/6$ trials at the
common load versus $3/6$ for the best baseline, FALCON, and pulls up to
$300$~kg compared with $200$~kg in the separate load sweep. In payload
carry, all controllers can physically carry the load, but only HULK and
AMO consistently remain within the prescribed path-deviation threshold,
both achieving $4/4$ successful trials at the common load. HULK and AMO
also both reach $12.25$~kg in the separate load sweep.  

The largest completion-rate gains arise when objects are supported
against the torso. HULK completes $4/4$ trials with both the fridge in
front of the torso and the canister at its side. AMO completes $2/4$
fridge trials, whereas Thor and FALCON complete none; all baselines
complete $0/4$ canister trials. HULK's maximum completed loads are
$7.0$~kg for the fridge and $6.1$~kg for the canister, compared with
baseline maxima of $6.0$ and $2.6$~kg, respectively. Together, these
results show that the distilled policy combines cart and payload
capabilities with successful whole-body carrying across the two tested
load geometries.


\begin{table*}[t]
    \centering
    \small
    \setlength{\tabcolsep}{4pt}
    \begin{tabular}{lcccccc}
        \toprule
        & \multicolumn{3}{c}{Tracking at $10$~kg/arm $\downarrow$} & \multicolumn{3}{c}{Disturbance rejection} \\
        \cmidrule(lr){2-4}\cmidrule(lr){5-7}
        Method & $E_{v_x}$ (m/s) & $E_{v_y}$ (m/s) & $E_{\omega_z}$ (rad/s) & \makecell{Worst DCM\\margin (m) $\uparrow$} & \makecell{Peak tilt\\(deg) $\downarrow$} & \makecell{Max push\\(N) $\uparrow$} \\
        \midrule
        Thor~\cite{thor2025} & $0.263 \pm 0.002$ & $0.265 \pm 0.003$ & $0.450 \pm 0.006$ & $-0.275 \pm 0.016$ & $23.17 \pm 0.13$ & $80$ \\
        AMO~\cite{amo} & $0.276 \pm 0.002$ & $0.350 \pm 0.004$ & $\mathbf{0.196 \pm 0.003}$ & $-0.705 \pm 0.030$ & $9.94 \pm 0.74$ & $50$ \\
        FALCON~\cite{falcon2026} & $0.367 \pm 0.012$ & $0.369 \pm 0.013$ & $0.392 \pm 0.005$ & $-0.329 \pm 0.034$ & $7.91 \pm 0.23$ & $80$ \\
        Ours & $0.242 \pm 0.009$ & $0.257 \pm 0.004$ & $0.234 \pm 0.008$ & $-0.251 \pm 0.011$ & $5.90 \pm 0.18$ & $\mathbf{110}$ \\
        Ours + CP-CBF & $\mathbf{0.210 \pm 0.007}$ & $\mathbf{0.142 \pm 0.003}$ & $ \mathbf{0.197 \pm 0.009}$ & $\mathbf{-0.162 \pm 0.012}$ & $\mathbf{5.59 \pm 0.24}$ & $\mathbf{130}$ \\
        \bottomrule
    \end{tabular}
    \caption{\textbf{Carrying and disturbance rejection.} Tracking under
    $10$~kg/arm at mid-forearm and, separately, standing torso pushes.
    Both Ours rows evaluate wrist-force teachers.
    DCM margin entries average episode-wise minimum signed margins; negative values
    indicate excursions outside the benchmark support envelope. Tilt is
    measured at $50$~N.
    Protocols, aggregation, and boldface are defined in \cref{sec:exp-setup}.}
    \label{tab:carrying_load}
\end{table*}

\begin{table*}[t]
    \centering
    \small
    \setlength{\tabcolsep}{4pt}
    \begin{tabular}{lcccccc}
        \toprule
        & \multicolumn{3}{c}{$E_{\mathrm{upper}}$ (rad) $\downarrow$} & \multicolumn{3}{c}{$E_{\mathrm{root}}$ (composite) $\downarrow$} \\
        \cmidrule(lr){2-4}\cmidrule(lr){5-7}
        Method & $\alpha_g\!=\!0$ & $0.5$ & $1.0$ & $\alpha_g\!=\!0$ & $0.5$ & $1.0$ \\
        \midrule
        Thor~\cite{thor2025} & $0.337 \pm 0.282$ & $0.329 \pm 0.282$ & $0.337 \pm 0.275$ & $0.762 \pm 0.264$ & $0.849 \pm 0.250$ & $0.913 \pm 0.185$ \\
        AMO~\cite{amo} & $0.402 \pm 0.207$ & $0.416 \pm 0.212$ & $0.445 \pm 0.222$ & $0.680 \pm 0.016$ & $1.228 \pm 0.064$ & $2.111 \pm 0.104$ \\
        FALCON~\cite{falcon2026} & $\mathbf{0.144} \pm 0.072$ & $\mathbf{0.148} \pm 0.074$ & $\mathbf{0.159} \pm 0.079$ & $\mathbf{0.444} \pm 0.041$ & $\mathbf{0.525} \pm 0.050$ & $0.868 \pm 0.134$ \\
        Ours & $0.161 \pm 0.052$ & $0.160 \pm 0.056$ & $0.162 \pm 0.059$ & $0.620 \pm 0.062$ & $0.543 \pm 0.044$ & $\mathbf{0.839} \pm 0.030$ \\
        Ours + CP-CBF & $0.172 \pm 0.055$ & $0.172 \pm 0.059$ & $0.176 \pm 0.061$ & $0.620 \pm 0.012$ & $0.752 \pm 0.019$ & $0.948 \pm 0.033$ \\
        \bottomrule
    \end{tabular}
    \caption{\textbf{Force rejection while walking.} Arm and composite root
    tracking errors; both Ours rows evaluate wrist-force teachers. Force scales
    $\alpha_g\in\{0,0.5,1.0\}$ correspond to $0$, $20$, and $40$~N per wrist. Metrics and reporting follow \cref{sec:exp-setup}.}
    \label{tab:hand_rejection_walking}
\end{table*}

\begin{table*}[t]
    \centering\small
    \setlength{\tabcolsep}{5pt}
    \begin{tabular}{lcccccccc}
        \toprule
        & \multicolumn{2}{c}{Cart push/pull} & \multicolumn{2}{c}{Payload carry}
        & \multicolumn{2}{c}{Fridge carry} & \multicolumn{2}{c}{Canister carry} \\
        \cmidrule(lr){2-3}\cmidrule(lr){4-5}\cmidrule(lr){6-7}\cmidrule(lr){8-9}
        Method & SR. $\uparrow$ & Max $\uparrow$ & SR. $\uparrow$ & Max $\uparrow$
               & SR. $\uparrow$ & Max $\uparrow$ & SR. $\uparrow$ & Max $\uparrow$ \\
        \midrule
        Thor~\cite{thor2025}     & 2/6 & 130 & 2/4 & 7.25 & 0/4 & 2.0 & 0/4  & 0.6 \\
        AMO~\cite{amo}           & 0/6 & 75 & \bf{4/4} & \bf{12.25} & 2/4 & 6.0 & 0/4  & 2.6 \\
        FALCON~\cite{falcon2026} & 3/6 & 200 & 0/4 & 3.62 & 0/4 & 2.5 & 0/4  & 0.0 \\
        \textbf{HULK}            & \bf{5/6}& \bf{300} & \bf{4/4} & \bf{12.25} & \bf{4/4} & \bf{7.0} & \bf{4/4}  & \bf{6.1} \\
        \bottomrule
    \end{tabular} 
    \caption{\textbf{Real-world evaluation on the Unitree G1.}
    \emph{SR.} reports successful trials/total trials at a common load per
    task. \emph{Max} reports the largest load completed in a separate
    sweep in kg.
    The HULK row evaluates the distilled student; see \cref{sec:exp-setup}
    for the protocol.}
    \label{tab:real}
    \vspace{-6pt}
\end{table*}

\section{Discussion and Limitations}

Load geometry helps explain why forceful loco-manipulation requires
whole-body coordination. A force's point of application determines its
moment about the CoM, coupling arm configuration to the support needed
to sustain the load (\cref{eq:affine}). Force-aware MPC exposes this
coupling in the training objective. Transfer from wrist forces to
forearm loading suggests that learned compensation can extend beyond
the trained contact location, although it does not establish
generalization to arbitrary contact arrangements.

MPC and CP-CBF provide complementary guidance: MPC supplies references
consistent with loaded dynamics, while the barrier shapes how the policy
approaches a balance boundary. The observed tradeoff between root
tracking and recovery (\cref{tab:carrying_load,tab:hand_rejection_walking})
suggests that carrying may benefit from allowing motion errors to preserve
balance. The exported policy is unfiltered, so the barrier provides no
runtime safety guarantee.

The model and training regime limit the behaviors this guidance can
support. The centroidal predictor approximates articulated dynamics
without explicitly resolving joint-level actuation feasibility, so a
balanced centroidal plan need not be executable by every joint.
The wrist-force teacher samples static arm poses, and the grasp teacher
starts from established grasps generated across multiple objects.
Coverage of stable configurations does not establish competence during
transitions between them.
Grasp acquisition, regrasping, and rapid contact changes therefore remain
outside the demonstrated scope. Richer dynamics models and training on
contact transitions are directions for extending to dynamic manipulation.

\section*{ACKNOWLEDGMENT}

We acknowledge the use of AI-generated content (including but not limited to text, figures, images, and code) in this paper.

\bibliographystyle{IEEEtran}
\bibliography{references}

\clearpage

\appendix

\makeatletter
\setlength{\@fptop}{0pt}
\setlength{\@fpsep}{12pt}
\setlength{\@fpbot}{0pt plus 1fil}
\setlength{\@dblfptop}{0pt}
\setlength{\@dblfpsep}{12pt}
\setlength{\@dblfpbot}{0pt plus 1fil}
\makeatother
\setlength{\textfloatsep}{12pt plus 2pt minus 2pt}
\setlength{\dbltextfloatsep}{12pt plus 2pt minus 2pt}
\renewcommand{\textfraction}{0.05}

\subsection{Capture Point Derivation}
\label{app:cbf}

Starting from the linear inverted pendulum model (LIPM), we assume
$\mathbf{c}_{xy}$ is the $xy$ position of the robot CoM and its
$z$-coordinate is held constant. We denote $\mathbf{p} \in \mathbb{R}^2$
as the center of pressure (CoP) of the foot contacts with the ground
plane. This gives the following dynamics:
\begin{align*}
    \ddot{\mathbf{c}}_{xy}
    = \frac{g}{z}(\mathbf{c}_{xy}-\mathbf{p}),
\end{align*}
where $g$ is the gravitational acceleration and $z$ is the constant height
of the robot CoM. In our teacher training, we set $z$ to the commanded
CoM height. If we define $\omega_0 = \sqrt{g/z}$ as the natural
frequency of the LIPM, then we get
\begin{align*}
    \ddot{\mathbf{c}}_{xy}
    = \omega_0^2(\mathbf{c}_{xy}-\mathbf{p}).
\end{align*}

If we assume a fixed $\mathbf{p}=\mathbf{p}_0$ and set
$\mathbf{y} = \mathbf{c}_{xy} - \mathbf{p}_0$, then we get
\begin{align*}
    \ddot{\mathbf{y}} = \omega_0^2 \mathbf{y},
\end{align*}
which is a linear system with solution
\begin{align*}
    \mathbf{y}(t) &= \frac{1}{2}
    \left(
        \mathbf{c}_0 - \mathbf{p}_0
        + \frac{\dot{\mathbf{c}}_0}{\omega_0}
    \right)e^{\omega_0t} \\
    &\quad
    + \frac{1}{2}
    \left(
        \mathbf{c}_0 - \mathbf{p}_0
        - \frac{\dot{\mathbf{c}}_0}{\omega_0}
    \right)e^{-\omega_0t}.
\end{align*}
Here, $\mathbf{c}_0 = \mathbf{c}_{xy}(0)$,
$\mathbf{p}_0 = \mathbf{p}(0)$, and
$\dot{\mathbf{c}}_0 = \dot{\mathbf{c}}_{xy}(0)$. As
$t \rightarrow \infty$, the second term approaches zero while the first
term diverges. Setting the coefficient of the divergent term to zero,
\begin{align*}
    \mathbf{c}_0 - \mathbf{p}_0
    + \frac{\dot{\mathbf{c}}_0}{\omega_0} = 0,
\end{align*}
gives
\begin{align*}
    \mathbf{p}_0
    = \mathbf{c}_0 + \frac{\dot{\mathbf{c}}_0}{\omega_0},
\end{align*}
which is the capture point solution at $t=0$. We therefore define
\begin{align*}
    \boldsymbol{\xi}
    = \mathbf{c}_{xy}
    + \frac{\dot{\mathbf{c}}_{xy}}{\omega_0}
\end{align*}
as the instantaneous capture point, where $\boldsymbol{\xi}$ is the CoP
that would capture the current state. Under the constant-height LIPM, this quantity is also the planar
divergent component of motion (DCM), whose margins and excursions
are reported in \cref{sec:sim-eval}. Differentiating with respect
to time gives
\begin{align*}
    \dot{\boldsymbol{\xi}}
    = \dot{\mathbf{c}}_{xy}
    + \frac{\ddot{\mathbf{c}}_{xy}}{\omega_0}.
\end{align*}
Substituting the LIPM dynamics gives
\begin{align*}
    \dot{\boldsymbol{\xi}}
    &= \dot{\mathbf{c}}_{xy}
    + \omega_0(\mathbf{c}_{xy}-\mathbf{p}) \\
    &= \omega_0(\boldsymbol{\xi}-\mathbf{p}),
\end{align*}
showing that the capture point follows the divergent dynamics of the
LIPM relative to $\mathbf{p}$.

Driving $\boldsymbol{\xi}-\mathbf{p}$ to zero eliminates the divergent
component of motion, leaving the convergent LIPM dynamics under which
$\mathbf{c}_{xy}\rightarrow\mathbf{p}$. The robot can regulate
$\mathbf{p}$ by redistributing ground reaction forces (GRFs) across its
foot contacts, including through ankle torques that shift pressure within
a supporting foot (the ankle strategy) and load transfer between
supporting feet. However, the set of achievable CoP
locations is constrained by the support polygon $\mathcal{S}_{\mathrm{support}}$. Thus,
when $\boldsymbol{\xi}\in\mathcal{S}_{\mathrm{support}}$, there exists an admissible CoP
that can be driven toward the capture point to arrest the divergent
motion. Conversely, when $\boldsymbol{\xi}$ lies outside
$\mathcal{S}_{\mathrm{support}}$, no CoP achievable under the current support configuration
can coincide with the capture point.

This capture-point criterion describes stopping without stepping under
the LIPM assumptions. Because the capture point includes CoM velocity,
it also applies to dynamic walking, where subsequent foot placements
change the support region. A capture point outside the current support
polygon therefore does not by itself imply a fall.

This motivates using the distance between the capture point and the
feasible support region in Eq.~\ref{eq:hcp}. Keeping this distance
bounded maintains the robot near states in which its GRFs retain the
ability to modulate the CoP toward the capture point and arrest divergent
CoM motion. We next describe how CBF-RL uses this safety function to
guide the wrist-force teacher during training.

\subsection{CBF-RL Training and Reward Guidance}
\label{app:cbf-training}

\noindent\textbf{Safety filtering.}
The safe set is
$\mathcal{S}_{\mathrm{safe}}=\{\mathbf{s}\in\mathcal{S}:h_{\mathrm{cp}}(\mathbf{s})\geq0\}$.
For a valid CBF, continuously enforcing
$\dot h_{\mathrm{cp}}+\kappa_{\mathrm{CBF}}h_{\mathrm{cp}}\geq0$,
with $\kappa_{\mathrm{CBF}}>0$, gives
$h_{\mathrm{cp}}(t)\geq e^{-\kappa_{\mathrm{CBF}}t}h_{\mathrm{cp}}(0)$:
a trajectory starting in the safe set stays there, provided the
constraint remains feasible under the modeled dynamics~\cite{cbfqp2017,cbftheory2019}.

Our training filter uses a local approximation of this condition.
Let $\mathbf{q}_{\ell}$ be the measured lower-body joint positions and
$\mathbf{q}_{\ell}^{\pi}$ the joint targets generated by the policy action
$\mathbf{a}_t$. With local barrier sensitivity $\mathbf{g}_{\mathrm{cp}}$
and per-step gain $\alpha_{\mathrm{CBF}}=\Delta t\,\kappa_{\mathrm{CBF}}$,
the CBF-QP finds the closest target satisfying the constraint:
\begin{equation}
\begin{aligned}
    \mathbf{q}_{\ell}^{\mathrm{QP}}
    = \arg\min_{\mathbf{q}_{\ell}^{\mathrm{tgt}}}\quad&
    \tfrac{1}{2}\|\mathbf{q}_{\ell}^{\mathrm{tgt}}-\mathbf{q}_{\ell}^{\pi}\|_2^2 \\
    \text{s.t.}\quad&
    \mathbf{g}_{\mathrm{cp}}^\top
    (\mathbf{q}_{\ell}^{\mathrm{tgt}}-\mathbf{q}_{\ell})
    +\alpha_{\mathrm{CBF}}h_{\mathrm{cp}}\geq0.
\end{aligned}
\label{eq:app-cbf-qp}
\end{equation}
An admissible proposal passes through unchanged. For nonzero sensitivity,
a violating proposal is projected onto the constraint boundary in closed form,
avoiding an iterative optimization solver in the training loop.

\noindent\textbf{Reward guidance.}
During training, the simulator executes the filtered target
$\widetilde{\mathbf{q}}_{\ell}$, while the policy is optimized using its
original sampled action. Following CBF-RL~\cite{cbfrl2026}, we add
\begin{align}
    r_{\mathrm{CBF}}
      =\Delta t\,w_{\mathrm{CBF}}
      \Big(&\min(\bar\rho_{\mathrm{cp}},0) \nonumber\\
      &+\exp\!\left[
          -\frac{\|\widetilde{\mathbf{q}}_{\ell}-\mathbf{q}_{\ell}^{\pi}\|_2^2}
                {\sigma_{\mathrm{CBF}}^2}
      \right]-1\Big),
    \label{eq:app-cbf-reward}
\end{align}
where the proposed barrier residual used for the reward is
\begin{equation}
\begin{aligned}
    \bar\rho_{\mathrm{cp}}
      ={}&\mathbf{g}_{\mathrm{cp}}^\top
          (\mathbf{q}_{\ell}^{\pi}-\mathbf{q}_{\ell}) \\
      &+\alpha_{\mathrm{CBF}}\max(h_{\mathrm{cp}},h_{\min}).
\end{aligned}
\label{eq:app-cbf-reward-residual}
\end{equation}
The first term penalizes the proposed violation; the second penalizes
the correction. This encourages the policy to need less intervention.
The filter and auxiliary reward are removed at deployment.

\noindent\textbf{Implementation.}
We approximate $\mathbf{g}_{\mathrm{cp}}$ by
$\mathbf{J}_{\mathrm{CoM},xy}^{\top}\nabla_{\boldsymbol{\xi}_f}h_{\mathrm{cp}}$,
using the horizontal CoM Jacobian restricted to lower-body joints and
differentiating only the CoM-position contribution to the capture point.
We cap the correction norm from the closed-form QP solution at
$0.30\max(\|\mathbf{q}_{\ell}^{\pi}-\mathbf{q}_{\ell}\|_2,10^{-3}\,\mathrm{rad})$
to obtain $\widetilde{\mathbf{q}}_{\ell}$ and regularize the sensitivity
norm near zero. The reward uses $w_{\mathrm{CBF}}=150$,
$\sigma_{\mathrm{CBF}}=0.5$~rad, and $h_{\min}=-1$; the QP uses the
unclipped barrier value. The approximate, capped training filter and the
unfiltered deployed policy do not inherit the ideal invariance guarantee.
Section~\ref{sec:sim-eval} empirically evaluates the learned wrist-force
teachers' tracking performance and disturbance recovery.

\subsection{Wrist-Force Teacher MPC Optimization}
\label{app:wrist-mpc}

We specialize \cref{eq:mpc,eq:affine} to two foot wrenches and two
wrist forces, with state
$\mathbf{x}=(\mathbf{c},\boldsymbol{\Theta},\mathbf{l},
\boldsymbol{\kappa})$ and the input
\begin{equation}
    \mathbf{u}_k =
    (\mathbf{f}_{L,k},\boldsymbol{\tau}_{L,k},
     \mathbf{f}_{R,k},\boldsymbol{\tau}_{R,k},
     \mathbf{f}^{h}_{L,k},\mathbf{f}^{h}_{R,k})
     \in\mathbb{R}^{18}.
    \label{eq:app-wrist-input}
\end{equation}
All forces act on the robot in the world frame;
$\mathbf{d}_h(t)$ is the prescribed force applied at wrist $h$, and
$\mathbf{f}^{h}_{h,k}$ is the corresponding MPC decision variable.
These prescribed forces represent external loading without explicitly
modeling a payload's mass distribution or inertial dynamics.

For wrist sites $\mathcal{H}=\{L,R\}$ at time-varying positions
$\mathbf{p}_h(t)$, the perturbed linear and angular momenta
$\hat{\mathbf{l}}$ and $\hat{\boldsymbol{\kappa}}$ satisfy
\begin{align}
    \dot{\hat{\mathbf{l}}}(t)
      &=\dot{\mathbf{l}}(t)
        +\sum_{h\in\mathcal{H}}\mathbf{d}_h(t),
      \label{eq:app-wrist-linear-momentum}\\
    \dot{\hat{\boldsymbol{\kappa}}}(t)
      &=\dot{\boldsymbol{\kappa}}(t)
        +\sum_{h\in\mathcal{H}}
        \big(\mathbf{p}_h(t)-\mathbf{c}(t)\big)
            \times\mathbf{d}_h(t).
      \label{eq:app-wrist-angular-momentum}
\end{align}
where the unhatted rates use the same robot state and foot wrenches
without wrist perturbations.

The prediction uses application points from forward kinematics (FK)
of the retargeted arm command:
\begin{equation}
    \mathbf{p}^{\mathrm{cmd}}_h(t)
      = \mathbf{p}_T(t)+\mathbf{R}_T(t)\,
        \operatorname{FK}_h
        \big(\mathbf{q}^{\mathrm{cmd}}_{\mathrm{arm}}(t)\big),
    \label{eq:app-wrist-position}
\end{equation}
where $(\mathbf{p}_T,\mathbf{R}_T)$ is the measured torso pose and
$\operatorname{FK}_h$ returns torso-frame positions. Motion-frame
sampling and blending follow \cref{app:wrist-teacher-training}.

At solve time $t_s$, \cref{eq:mpc} uses the costs in
\cref{eq:stagecost,eq:termcost} and input reference
\begin{equation}
    \mathbf{u}^{\mathrm{ref}}_k
      =(\mathbf{0}_{12},\mathbf{d}_L(t_s),\mathbf{d}_R(t_s)).
    \label{eq:app-wrist-reference}
\end{equation}

The guidance reward in \cref{eq:mpcreward} tracks
\begin{align}
    \Lambda_{\mathrm{CoM}}(\mathbf{x},\mathbf{u})
      &=\mathbf{c}, &
    \Lambda_{\mathrm{CoMvel}}(\mathbf{x},\mathbf{u})
      &=\mathbf{l}/m, \nonumber\\
    \Lambda_{\mathrm{GRF},i}(\mathbf{x},\mathbf{u})
      &=\mathbf{f}_i, &
    \Lambda_{\mathrm{resist},h}(\mathbf{x},\mathbf{u})
      &=\mathbf{f}^{h}_h.
    \label{eq:app-wrist-lambda}
\end{align}

\subsection{Grasp Teacher MPC Optimization}
\label{app:grasp-mpc}

Across the training object set, the grasp teacher allocates contact
wrenches at each established torso-and-arm grasp to balance the assigned
object, then supplies their reaction to the locomotion MPC.

\noindent\textbf{Object equilibrium.}
For the object assigned to each environment, let $\mathcal{G}$ be the
active grasp contacts, $\mathbf{p}_j$ their world-frame positions, and
$(m_o,\mathbf{c}_o)$ its mass and CoM.
Define $\mathbf{g}_j$ and $\boldsymbol{\nu}_j$ as the force and contact
moment exerted by the robot \emph{on the object} at contact $j$.
Under the quasi-static object model, equilibrium requires
\begin{align}
    \sum_{j\in\mathcal{G}}\mathbf{g}_j+m_o\mathbf{g}
      &=\mathbf{0}, \label{eq:app-grasp-force-balance}\\
    \sum_{j\in\mathcal{G}}
      \big[(\mathbf{p}_j-\mathbf{c}_o)\times\mathbf{g}_j
                +\boldsymbol{\nu}_j\big]
      &=\mathbf{0}. \label{eq:app-grasp-moment-balance}
\end{align}

The coupled friction constraint is
\begin{equation}
    \|\boldsymbol{\eta}_j\|_2\leq g^n_j,\qquad
    0\leq g^n_j\leq g^{n,\max}_j,
    \label{eq:app-grasp-wrench-cone}
\end{equation}
where $g^n_j$ is the inward normal force, $g^{n,\max}_j$ its cap, and
$\boldsymbol{\eta}_j$
stacks the two tangential-force and three contact-moment components
in the contact frame, divided by their respective sliding, torsional,
and rolling friction coefficients. Zero-friction components are fixed
to zero and omitted. We then apply these forces to the centroidal dynamics and integrate the equilibrium constraints into the MPC for the contact force decision variables to solve.

\noindent\textbf{Grasp-force guidance through $\Lambda$.}
Let $\mathbf{z}^{g}$ collect grasp wrenches and $\mathbf{u}^{f}$ foot wrenches.
The scale $\ell_\tau>0$ converts moments to force units.
The CoM and GRF maps in \cref{eq:app-wrist-lambda} are retained, while
the grasp output for each torso or arm body $b$ is
\begin{equation}
    \Lambda_{\mathrm{grasp},b}
       (\mathbf{x},\mathbf{u}^{f},\mathbf{z}^{g})
       =(\mathbf{G}_b,\mathbf{N}_b/\ell_\tau),
    \label{eq:app-grasp-lambda}
\end{equation}
where $\mathbf{G}_b=\sum_{j\in\mathcal{G}_b}\mathbf{g}_j$ and
$\mathbf{N}_b=\sum_{j\in\mathcal{G}_b}
[(\mathbf{p}_j-\mathbf{p}_b)\times\mathbf{g}_j+\boldsymbol{\nu}_j]$,
with $\mathcal{G}_b$ the contacts on body $b$ and $\mathbf{p}_b$ the
common reference point for planned and measured wrenches.
Their tracking error and reward are
\begin{align}
    E_g&=\sum_b\|
       \tilde{\Lambda}_{\mathrm{grasp},b}
       -\Lambda^{\mathrm{meas}}_{\mathrm{grasp},b}\|_2^2,
       \nonumber\\
    r_{\mathrm{grasp}}&=
       \begin{cases}
         (1+E_g/\sigma_g^2)^{-1},&\text{valid allocation},\\
         0,&\text{otherwise},
       \end{cases}
       \label{eq:app-grasp-tracking}
\end{align}
with $\sigma_g=25$~N and tilde denoting the plan. Validity requires
active contact, finite wrenches, and solver convergence with
\cref{eq:app-grasp-force-balance,eq:app-grasp-moment-balance,eq:app-grasp-wrench-cone}
satisfied within numerical tolerances.
Otherwise, measured contact reactions supply the locomotion MPC load.

\subsection{Wrist-Force Teacher Training}
\label{app:wrist-teacher-training}

\noindent\textbf{Upper-body command curriculum.}
We uniformly sample individual frames, not sequences, from $48{,}233$
retargeted frames across $252$ motions, retaining $14$ arm joint angles.
For environment $e$, coverage $c_e\in[0.3,1]$ contracts the sampled
pose toward the nominal pose:
\begin{equation}
    \mathbf q^{\mathrm{ref}}_e
      =\mathbf q^0+c_e(\mathbf q^{\mathrm{sample}}_e-\mathbf q^0).
    \label{eq:app-arm-coverage}
\end{equation}
Joint-limited targets resample every $3$--$8$~s, with a $1$~s linear
blend from the measured pose followed by a hold.

Starting at $c_e=0.3$, coverage updates at reset using the completed
episode length $L_e$ in policy steps:
\begin{equation}
    c_e^+=\operatorname{clip}_{[0.3,1]}
      \left(c_e+0.01\,\mathbf1[L_e>200]
                  -0.01\,\mathbf1[L_e<40]\right).
    \label{eq:app-arm-curriculum}
\end{equation}
At $50$~Hz the thresholds are $4$~s and $0.8$~s.

\noindent\textbf{Force curriculum.}
Each force-active environment independently updates a magnitude factor
$f_e\in[0,1]$, initialized to zero, at reset:
\begin{equation}
    f_e^+=\operatorname{clip}_{[0,1]}
      \left(f_e+0.02\,\mathbf1[L_e>200]
                  -0.02\,\mathbf1[L_e<200]\right).
    \label{eq:app-force-curriculum}
\end{equation}
A fixed $20\%$ subset remains force-free with no curriculum advancement,
preserving exposure to unloaded conditions throughout training and
discouraging reliance on external forces for balance and command tracking.

For each side, heading-frame components are sampled independently from
$[-40,40]$, $[-40,40]$, and $[-50,5]$~N, scaled by $f_e$, independently
zeroed with probability $0.25$, and held for $3$--$5$~s.
During walking, horizontal forces oppose the planar velocity command:
\begin{equation}
    \mathbf u=-\frac{\mathbf v^{\mathrm{cmd}}_{xy}}
                        {\|\mathbf v^{\mathrm{cmd}}_{xy}\|_2},
    \qquad
    \mathbf F_{xy}\leftarrow
         |\mathbf F_{xy}^{\top}\mathbf u|\,\mathbf u.
    \label{eq:app-force-direction}
\end{equation}
Standing retains the sampled direction; vertical forces are unchanged.
Current base yaw rotates the forces into the world frame.


\subsection{Grasp Teacher Training}
\label{app:grasp-teacher-training}

\begin{figure}[t]
    \centering
    \includegraphics[width=\columnwidth]{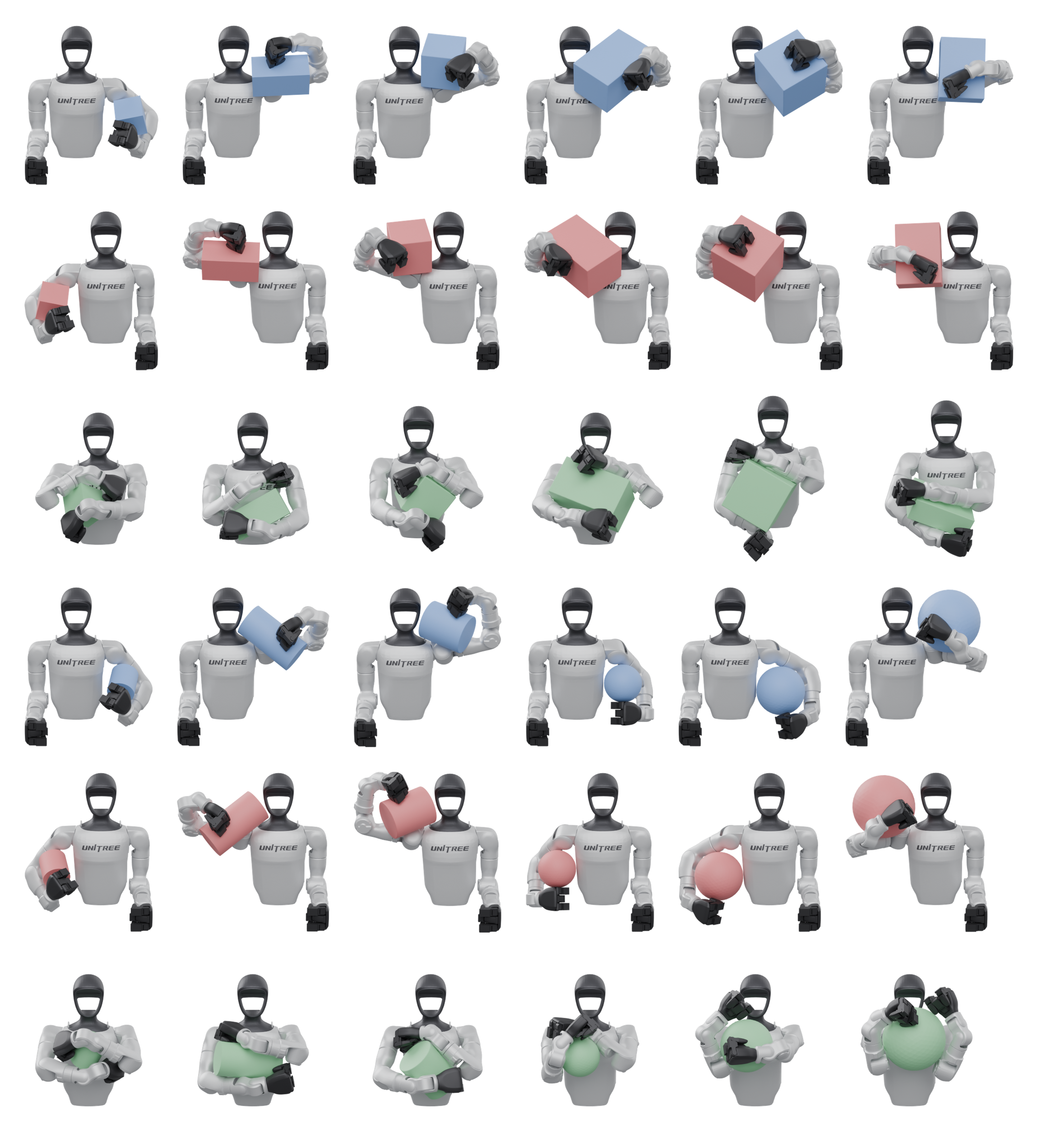}
    \caption{\textbf{Example whole-body grasp configurations.}
    Authored torso-and-arm postures and object transforms are restored
    together at reset. These illustrate initialization geometry, not
    cross-object policy evaluation.}
    \label{fig:app-grasp-examples}
\end{figure}

\noindent\textbf{Initialization and commanded grasp.}
We initialize from a locomotion and static-arm checkpoint, not a
grasp-force-trained policy. In the initialization study, only actor
and critic network weights transfer; optimizer and curriculum state
start afresh.
The grasp bank spans multiple object geometries, with each environment
assigned an object and a compatible grasp. Reset restores the authored
arm posture and the assigned object's transform together, clears velocity,
and resets the relative object reference for slip and drop measurement
(\cref{fig:app-grasp-examples}).
Upper-body residuals about the grasp use a $0.05$~rad scale; lower-body
offsets use $0.5$~rad. Motion-bank coverage is disabled to preserve the
grasp geometry.

\noindent\textbf{Per-environment payload curriculum.}
With an assigned object--grasp pair per environment, each environment
maintains an independent upper mass bound $b_e$, initially $0.5$~kg.
At reset,
\begin{equation}
    m_e\sim\mathcal U(0.5,b_e),\qquad 0.5\leq b_e\leq5\ \mathrm{kg}.
    \label{eq:app-payload-sampling}
\end{equation}
A full $20$~s episode without terminal failure increases the bound;
a terminal failure decreases it:
\begin{equation}
    b_e^+=
    \begin{cases}
      \min(5,1.1b_e), & \text{full-episode success},\\
      \max(0.5,b_e/1.1), & \text{terminal failure},\\
      b_e, & \text{otherwise}.
    \end{cases}
    \label{eq:app-payload-update}
\end{equation}
Masses are in kilograms. Administrative resets and shortened timeouts
do not increase $b_e$. Mass stays fixed within each episode, with
inertia scaled proportionally at reset.
For resampled grasps, the bound follows the identity of the grasp
and its object across resets. Duplicate grasps in a reset batch update at
most once, with terminal failure taking precedence.


\subsection{Teacher Training Curves}
\label{app:teacher-training-curves}

\begin{figure*}[t]
    \centering
    \includegraphics[width=\textwidth]{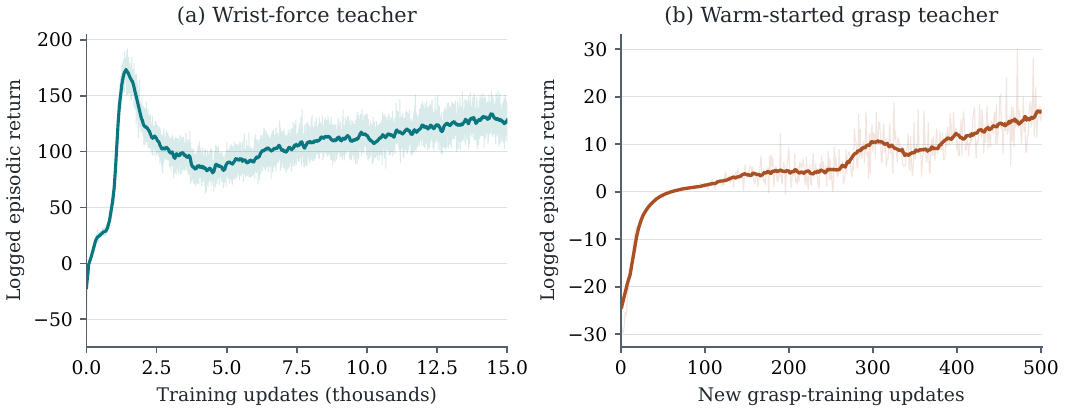}
    \caption{\textbf{Recorded teacher training returns.}
    Both panels show the reward training curve during training for both teachers. The plots use a moving average filter on the actual training curve to produce a cleaner plot, but the original curve is shown in a lighter color. For the grasp teacher, we do not include the pre-training steps.}
    \label{fig:app-teacher-returns}
\end{figure*}

\Cref{fig:app-teacher-returns} reports rank-zero rolling returns over
$100$ completed episodes for one run of each teacher, both using seed $42$.
The wrist-force run spans $15{,}000$ updates with $8$ synchronized ranks;
the grasp run spans $501$ new updates with $32$ synchronized ranks.
Both use $4096$ environments per rank and $24$ steps per update.
The grasp curve measures early adaptation, not convergence.

\subsection{Teacher MDPs}
\label{app:e1mdp}

The wrist-force and grasp teachers use separate Markov decision processes
(MDPs), with shared actor observations listed in \cref{tab:e1mdp}.
Their rewards combine command tracking, MPC guidance, and action and joint
regularization; the grasp teacher additionally rewards object retention
(\cref{app:rewards}). Critic networks receive privileged simulation observations
(\cref{tab:critic-observations}).

\begin{table}[t]
    \centering
    \footnotesize
    \begin{tabular}{lr}
        \toprule
        Observation & Values per sample \\
        \midrule
        Base angular velocity & $3$ \\
        Projected gravity & $3$ \\
        Joint position / velocity & $29$ / $29$ \\
        Previous action & $29$ \\
        Commanded planar velocity / yaw rate & $3$ \\
        Gait phase (sine and cosine) & $2$ \\
        Base-height command & $1$ \\
        Torso roll/pitch error & $2$ \\
        Arm joint target & $14$ \\
        \midrule
        Actor total, one sample & $115$ \\
        Actor total, five samples & $575$ \\
        \bottomrule
    \end{tabular}
    \caption{\textbf{Shared observations for wrist-force and grasp teacher actors.}
    Observation noise is uniform:
    angular velocity $\pm0.2$~rad/s, gravity components $\pm0.05$,
    joint position $\pm0.01$~rad, and joint velocity $\pm1.5$~rad/s.
    Privileged critic observations are listed in \cref{tab:critic-observations}.}
    \label{tab:e1mdp}
\end{table}


Domain randomization varies foot friction over $0.3$--$1.2$,
proportional/derivative gains by $\times0.8$--$1.2$, rotor inertia by
$\times0.5$--$1.5$, and base mass by $-1$ to $+3$~kg.
Base CoM offsets span $\pm0.025$~m horizontally and $\pm0.030$~m
vertically; pseudo-inertia parameter $\alpha\in[-0.112,0.091]$
and encoder bias spans $\pm0.015$~rad.
Action delay ramps to $20$~ms over $2500$ updates after a
$1500$-update warmup.

Wrist-force training uses the applied-force curriculum in
\cref{app:wrist-teacher-training}, with base-velocity perturbations
every $4$--$6$~s:
$v_x,v_y\in[-2,2]$~m/s, $v_z\in[-0.4,0.4]$~m/s,
roll/pitch rates within $\pm0.52$~rad/s, and yaw rate within
$\pm0.78$~rad/s. Sustained pushes are attempted at the same interval
with probability $0.5$ and last $0.5$--$1.2$~s; their magnitude
ramps to $150$~N over $4000$ updates after a $300$-update warmup.
Grasp training disables these perturbations and uses loads from multiple objects,
with initialization and payload sampling as in
\cref{app:grasp-teacher-training}. Its height and torso commands resample
every $5$~s, and action delay is sampled uniformly in simulation steps.
Both teachers terminate on base tilt beyond $70^\circ$ or non-foot
ground contact. Grasp episodes also terminate on invalid simulation
states or object drop:
relative object drift above $0.15$~m or height below $0.30$~m,
monitored from reset.

\begin{table}[t]
    \centering
    \small
    \begin{tabular}{lr}
        \toprule
        Critic-only observation & Values \\
        \midrule
        Base linear velocity & $3$ \\
        Foot height / air time / contact flag & $2+2+2$ \\
        Foot contact-force vectors & $6$ \\
        MPC CoM reference & $3$ \\
        MPC angular-momentum reference & $3$ \\
        MPC tilt reference & $3$ \\
        Filtered yaw rate & $1$ \\
        \midrule
        Additional privileged values & $25$ \\
        Actor channels, one current sample & $115$ \\
        Total per critic & $140$ \\
        \bottomrule
    \end{tabular}
    \caption{\textbf{Privileged observations for teacher critics.}
    Each critic receives $140$ current-state values without added noise.
    Actors instead stack five samples of $115$ values, spanning $80$~ms
    at $50$~Hz, without direct object-pose or contact-force observations.}
    \label{tab:critic-observations}
\end{table}

\begin{table}[t]
    \centering
    \small
    \setlength{\tabcolsep}{3pt}
    \begin{tabular}{lc}
        \toprule
        Hyperparameter & Value \\
        \midrule
        Rollout length per environment & $24$ steps ($0.48$~s) \\
        Samples per GPU ($4096$ environments) & $98304$ \\
        Learning epochs / minibatches & $5$ / $4$ \\
        Samples per minibatch per GPU & $24576$ \\
        Optimizer steps/partition/update & $20$ \\
        Discount $\gamma$ / GAE $\lambda$ & $0.99$ / $0.95$ \\
        PPO clipping ratio & $0.2$ \\
        Entropy / value-loss coefficient & $0.01$ / $1.0$ \\
        Clipped value loss & Enabled \\
        Per-minibatch advantage normalization & Disabled \\
        Initial learning rate & $10^{-3}$ \\
        Learning-rate schedule & Adaptive, per partition \\
        Target KL divergence & $0.01$ \\
        Maximum gradient norm & $1.0$ \\
        Optimizers per teacher & Two Adam optimizers \\
        Action standard deviation bounds & $[0.01,5]$ \\
        Action clipping & None \\
        Intrinsic reward & None \\
        \bottomrule
    \end{tabular}
    \caption{\textbf{Shared PPO settings.} Generalized advantage estimation
    (GAE) uses $\lambda=0.95$; the adaptive learning-rate schedule targets
    Kullback--Leibler (KL) divergence $0.01$ separately for each body partition.
    Batch counts are per graphics processing unit (GPU), each simulating
    $4096$ environments, with synchronized training across GPUs.
    The two body partitions share transitions but use separate rewards
    and updates.}
    \label{tab:ppo}
\end{table}

\subsection{Reward Formulation}
\label{app:rewards}

For body partition $g\in\{\mathrm{lower},\mathrm{upper}\}$, the reward is
$r_g=\Delta t\sum_{i\in\mathcal I_g}w_i f_i$, where $\mathcal I_g$
contains its reward terms, $f_i$ is the term value, $w_i$ its weight,
and $\Delta t=0.02$~s.
\Cref{tab:rewards,tab:carry-rewards} summarize the term formulations
and weights for both teachers; the barrier reward is defined in
\cref{app:cbf-training}.
We write $[x]_+=\max(x,0)$ and $\mathbf{1}[A]$ for the indicator of
condition $A$. Motion gates use
$s=\|\mathbf v_{xy}^{\mathrm{cmd}}\|_2+|\omega_z^{\mathrm{cmd}}|$;
gated terms are zero when inactive unless stated otherwise.
The vector $\hat{\mathbf g}^{\,B}$ is unit gravity in frame $B$, and
$\operatorname{wrap}$ maps angles to $[-\pi,\pi)$.

MPC errors $\mathbf e_{x,\ell}=\mathbf x-\mathbf x_\ell^{\mathrm{MPC}}$
use lookahead index $\ell$; $\dot{\boldsymbol{\kappa}}$ is estimated
by finite differences. For foot $i$, $c_i$, $c_i^{\mathrm{gait}}$, and
$d_i$ indicate measured contact, scheduled contact, and first touchdown;
$h_i,h_i^{\mathrm{peak}}$ are sole and peak swing heights, and
$\mathbf v_i,\mathbf F_i$ are velocity and contact force.
Lengths and times are in meters and seconds.



\begin{table*}[t]
    \centering
    \footnotesize
    \renewcommand{\arraystretch}{1.15}
    \setlength{\tabcolsep}{3pt}
    \begin{tabular*}{\textwidth}{@{\extracolsep{\fill}}>{\raggedright\arraybackslash}p{0.19\textwidth}>{\raggedright\arraybackslash}p{0.59\textwidth}rr@{}}
        \toprule
        Term & Formulation & Wrist-force & Grasp \\
        \midrule
        \multicolumn{4}{l}{\emph{MPC guidance}} \\
        CoM position & $\exp[-\sum_{\ell=1}^{5}\alpha_\ell^c\|\mathbf e_{c,\ell}\|_2^2]$. & $2$ & $2$ \\ 
        CoM velocity & $\exp[-8\sum_{\ell=1}^{3}\alpha_\ell^v\|\mathbf e_{v,\ell}\|_2^2]$. & $4$ & $4$ \\ 
        Angular momentum & $-\sum_{\ell=1}^{5}\alpha_\ell^c(\|\mathbf e_{\kappa,\ell}\|_2^2+0.01\,\mathbf e_{\dot\kappa,\ell}^{\top}\mathbf{Q}_{\dot\kappa}\mathbf e_{\dot\kappa,\ell})$. & $0.01$ & $0.01$ \\ 
        Ground reaction forces & $\exp[-\frac{1}{|\mathcal V|}\sum_{i\in\mathcal V}\|\mathbf e_{F,i}\|_2^2/(150\,\mathrm{N})^2]$; $\mathcal V$: valid loaded stance feet; $0$ if empty. & $0.002$ & $0.002$ \\ 
        Swing foothold & $\frac{1}{|\mathcal S|}\sum_{i\in\mathcal S}\exp[-\|\mathbf e_{p,i,xy}\|_2^2/(0.08\,\mathrm{m})^2]$; $\mathcal S$: valid swing feet; $1$ if empty or $s\leq0.1$. & $2$ & $2$ \\ 
        Base tilt & $\exp[-4\sum_{\ell=1}^{3}\alpha_\ell^\theta\|\mathbf e_{\theta,\ell}\|_2^2]$. & $2$ & $2$ \\ 
        External-force resistance$^\dagger$ & $\frac12\sum_{i=1}^{2}\exp[-\|\mathbf F_i^{e,\mathrm{applied}}-\mathbf F_i^{e,\mathrm{MPC}}\|_2^2/(30\,\mathrm{N})^2]$; inactive sides contribute $1$. & $1$ & --- \\ 
        \midrule
        \multicolumn{4}{l}{\emph{Command tracking}} \\
        Linear velocity & $\exp[-(\|\mathbf{v}_{xy}^{b}-\mathbf{v}_{xy}^{\mathrm{cmd}}\|_2^2+(v_z^b)^2)/0.5^2]$; base frame. & $0.5$ & $0.5$ \\ 
        Yaw rate & $\exp[-(\bar{\omega}_z-\omega_z^{\mathrm{cmd}})^2/0.2^2]$; $0.4$~s yaw-rate filter. & $3$ & $3$ \\ 
        Base height & $\exp(-|z-z^{\mathrm{cmd}}|/0.1)$. & $3$ & $3$ \\ 
        Torso roll/pitch & $\exp(-\|\boldsymbol{\theta}-\boldsymbol{\theta}^{\mathrm{cmd}}\|_2/0.15)$; $\boldsymbol{\theta}$: torso roll/pitch. & $3$ & $3$ \\ 
        Arm joint targets$^\dagger$ & $\exp[-\sum_{j\in\mathrm{arms}}(q_j-q_j^{\mathrm{cmd}})^2/0.5^2]$. & $4$ & $4$ \\ 
        \midrule
        \multicolumn{4}{l}{\emph{Gait and swing shaping}} \\
        Gait-phase contact match & $\frac12\sum_{i=1}^{2}\mathbf{1}[c_i=c_i^{\mathrm{gait}}]$, for $s>0.2$. & $4$ & $4$ \\ 
        Air time & $\sum_{i=1}^{2}\mathbf{1}[0.02<t_i^{\mathrm{air}}<0.45]$,     for $s>0.5$. & $4$ & $4$ \\ 
        Flat sole orientation & $\frac12\sum_{i=1}^{2}\exp[-2(Q_{i,x}^2+Q_{i,y}^2)/0.15]$; $\mathbf Q_i$: sole quaternion. & $1$ & $1$ \\ 
        Center of pressure & $\frac12\sum_{i=1}^{2}\exp[-\tilde e_i^2/0.25^2]$, where $\tilde e_i=(\rho_i-0.5)(1+\mathbf{1}[\rho_i<0.5])$ and $\rho_i$ is toe-load fraction; unloaded feet contribute one. & $1$ & $1$ \\ 
        Foot clearance & $\sum_{i=1}^{2}[|h_i-0.12|-0.02]_+$, for $s>0.2$. & $-5$ & $-5$ \\ 
        Sole swing height & $\sum_{i=1}^{2}d_i(h_i^{\mathrm{peak}}/0.12-1)^2$, for $s>0.2$. & $-20$ & $-20$ \\ 
        Foot slip & $\sum_{i=1}^{2}c_i\|\mathbf v_{i,xy}\|_2^2$, for $s>0.05$. & $-0.1$ & $-0.1$ \\ 
        Soft landing & $\sum_{i=1}^{2}d_i\|\mathbf F_i\|_2$, for $s>0.05$. & $-10^{-5}$ & $-10^{-5}$ \\ 
        Support-mode mismatch & $\mathbf{1}[n_c\neq1]$ for $s>0.2$, otherwise $\mathbf{1}[n_c<2]$, with $n_c=c_1+c_2$. & $-2$ & $-2$ \\ 
        Feet separation & $\max(0.12-\|\mathbf{p}_{L,xy}-\mathbf{p}_{R,xy}\|_2,0)$. & $-0.5$ & $-0.5$ \\ 
        Sole orientation & $\sum_{i=1}^{2}\|\hat{\mathbf g}^{\,\mathrm{sole}_i}_{xy}\|_2$. & $-2$ & $-2$ \\ 
        Ankle roll & $\sum_{j\in\mathrm{ankle\,roll}}|q_j|$. & $-2$ & $-2$ \\ 
        Feet heading & $\sum_{i=1}^{2}|\operatorname{wrap}(\psi_i-\psi_{\mathrm{base}}-0.4\,\omega_z^{\mathrm{cmd}})|$, where $\psi$ denotes heading. & $-1$ & $-1$ \\ 
        \bottomrule
    \end{tabular*}
    \caption{\textbf{Teacher tracking and gait rewards.}
    $\dagger$ marks upper-body terms; others enter the lower-body reward.
    A dash denotes an absent term.}
    \label{tab:rewards}
\end{table*}

\begin{table*}[t]
    \centering
    \footnotesize
    \renewcommand{\arraystretch}{1.15}
    \setlength{\tabcolsep}{3pt}
    \begin{tabular*}{\textwidth}{@{\extracolsep{\fill}}>{\raggedright\arraybackslash}p{0.19\textwidth}>{\raggedright\arraybackslash}p{0.59\textwidth}rr@{}}
        \toprule
        Term & Formulation & Wrist-force & Grasp \\
        \midrule
        \multicolumn{4}{l}{\emph{Posture and standstill}} \\
        Pelvis upright & $\exp[-5\|\hat{\mathbf g}^{\,\mathrm{pelvis}}_{xy}\|_2^2]$. & $1$ & $1$ \\ 
        Torso upright & $\exp[-5\|\hat{\mathbf g}^{\,\mathrm{torso}}_{xy}\|_2^2]$. & $0$ & $1$ \\ 
        Nominal posture & $\exp[-\operatorname{mean}_j((q_j-q_j^0)^2/\sigma_j^2)]$ over legs and waist, with speed-dependent joint tolerances. & $1$ & $1$ \\ 
        Base-height error & $(1+4\,\mathbf{1}[s\leq0.2])(z-z^{\mathrm{cmd}})^2$. & $-10$ & $-10$ \\ 
        Leg splay & $\sum_{j\in\mathrm{hip\,roll/yaw}}q_j^2$. & $-4$ & $-4$ \\ 
        Waist anchor & $(q_{\mathrm{waist,roll}}-\theta_{\mathrm{roll}}^{\mathrm{cmd}})^2+(q_{\mathrm{waist,pitch}}-\theta_{\mathrm{pitch}}^{\mathrm{cmd}})^2$. & $-1$ & $-1$ \\ 
        Waist yaw & $q_{\mathrm{waist,yaw}}^2$. & $-6$ & $-2$ \\ 
        Chest heading & $|\operatorname{wrap}(\psi_{\mathrm{torso}}-\psi_{\mathrm{base}})|$, where $\psi$ denotes heading. & $-6$ & $-2$ \\ 
        Waist effort & $\sum_{j\in\mathrm{waist}}[\max(|\tau_j|/\tau_j^{\max}-0.7,0)]^2$. & $-10$ & $-10$ \\ 
        Body angular velocity & $\omega_x^2+\omega_y^2$ for the pelvis angular velocity in the world frame; yaw is excluded. & $-0.05$ & $-0.05$ \\ 
        Standing joint deviation & $\sum_{j\in L\cup U}(q_j-q_j^0)^2$, when all task commands request rest. & $-1$ & $-1$ \\ 
        Standing base drift & $\|\mathbf v_{\mathrm{base},xy}\|_2^2$, when all task commands request rest. & $-6$ & $-6$ \\ 
        \midrule
        \multicolumn{4}{l}{\emph{Joint regularization}} \\
        All-action rate & $\|\mathbf{a}_t-\mathbf{a}_{t-1}\|_2^2$ over all raw joint actions. & $-0.25$ & $-0.25$ \\ 
        Upper action rate$^\dagger$ & $\|\mathbf{a}_t^{U}-\mathbf{a}_{t-1}^{U}\|_2^2$ over raw arm actions. & $-0.1$ & $-0.1$ \\ 
        All-joint limits & $\sum_{j\in L\cup U}([q_j^{\min}-q_j]_++[q_j-q_j^{\max}]_+)$. & $-1$ & $-1$ \\ 
        Upper joint limits$^\dagger$ & $\sum_{j\in U}([q_j^{\min}-q_j]_++[q_j-q_j^{\max}]_+)$. & $-1$ & $-1$ \\ 
        Lower joint velocity & $\sum_{j\in L}\dot q_j^2$. & $-2.5{\times}10^{-3}$ & $-2.5{\times}10^{-3}$ \\ 
        Upper joint velocity$^\dagger$ & $\sum_{j\in U}\dot q_j^2$. & $-10^{-3}$ & $-10^{-3}$ \\ 
        Lower joint acceleration & $\sum_{j\in L}\ddot q_j^2$. & $-7.5{\times}10^{-7}$ & $-7.5{\times}10^{-7}$ \\ 
        Upper joint acceleration$^\dagger$ & $\sum_{j\in U}\ddot q_j^2$. & $-1.2{\times}10^{-8}$ & $-1.2{\times}10^{-8}$ \\ 
        Lower joint torque & $\sum_{j\in L}\tau_j^2$. & $-8.75{\times}10^{-5}$ & $-8.75{\times}10^{-5}$ \\ 
        Upper joint torque$^\dagger$ & $\sum_{j\in U}\tau_j^2$. & $-3.5{\times}10^{-5}$ & $-3.5{\times}10^{-5}$ \\ 
        \midrule
        \multicolumn{4}{l}{\emph{Object support and retention}} \\
        Grasp-wrench tracking$^\dagger$ & $[1+E_g/(25\,\mathrm{N})^2]^{-1}$ for a valid allocation, otherwise zero; $E_g$ is the squared wrench error in \cref{eq:app-grasp-tracking}. & --- & $8$ \\ 
        Grasp-force regularization$^\dagger$ & $e/(e+50\,\mathrm{N})$, where $e=\max(F-30\,\mathrm{N},0)$ and $F$ sums measured contact-region force magnitudes. & --- & $-0.5$ \\ 
        Object drop, lower & $\mathbf{1}[\text{object-drop termination}]$ & --- & $-250$ \\ 
        Object drop, upper$^\dagger$ & $\mathbf{1}[\text{object-drop termination}]$ & --- & $-250$ \\ 
        Object slip, lower & $\|\mathbf v_{\mathrm{obj}/T}\|_2$, relative to a torso & --- & $-2$ \\ 
        Object slip, upper$^\dagger$ & $\|\mathbf v_{\mathrm{obj}/T}\|_2$, relative to a torso & --- & $-2$ \\ 
        Lower torque-limit excess & $\sum_{j\in L}\max(|\tau_j|-0.9\,\tau_j^{\max},0)$. & --- & $-0.1$ \\ 
        Upper torque-limit excess$^\dagger$ & $\sum_{j\in U}\max(|\tau_j|-0.9\,\tau_j^{\max},0)$. & --- & $-0.1$ \\ 
        \bottomrule
    \end{tabular*}
    \caption{\textbf{Teacher posture, regularization, and grasp rewards.}
    $L$: leg/waist joints; $U$: arm joints; $q_j^0$: default joint position;
    $q_j^{\min},q_j^{\max}$: soft limits.
    Markers follow \cref{tab:rewards}.}
    \label{tab:carry-rewards}
\end{table*}

\subsection{Teacher Dual-Policy Training}
\label{app:ppo}

Following \cite{falcon2026}, each teacher uses separate
lower-body and upper-body actor-critic pairs. Both actors receive the same
full-body history and commands, while both critics receive the same
privileged observations. Their Gaussian actions concatenate into
$15$ lower-body and $14$ upper-body joint offsets.
The pairs train on shared trajectories using separate reward streams
(\cref{tab:rewards,tab:carry-rewards}) and Adam optimizers under
Proximal Policy Optimization (PPO; \cref{tab:ppo}).

\end{document}